\documentclass[10pt]{article} 

\usepackage{amsmath,amssymb,amsthm}
\usepackage{algorithm}
\usepackage{algorithmic}
\usepackage{graphicx}
\usepackage{hyperref}
\usepackage{enumerate}
\usepackage{natbib}
\usepackage{color}
\usepackage{mathtools}
\usepackage{bm}
\usepackage{bbm}
\usepackage{natbib}

\usepackage[margin=1.5in]{geometry}
\usepackage{hyperref}

\newcommand{\R}{\mathbb R}

\newcommand{\N}{\mathbb N}
\newcommand{\E}{\mathbb E}

\def\bA{\boldsymbol{A}}
\def\bB{\boldsymbol{B}}
\def\bC{\boldsymbol{C}}

\def\bG{\boldsymbol{G}}

\def\bI{\boldsymbol{I}}

\def\bK{\boldsymbol{K}}

\def\bM{\boldsymbol{M}}

\def\bP{\boldsymbol{P}}
\def\bQ{\boldsymbol{Q}}

\def\bS{\boldsymbol{S}}

\def\bU{\boldsymbol{U}}
\def\bV{\boldsymbol{V}}

\def\bX{\boldsymbol{X}}
\def\bY{\boldsymbol{Y}}

\def\cA{\mathcal{A}}
\def\cB{\mathcal{B}}

\def\cH{\mathcal{H}}

\def\cK{\mathcal{K}}

\def\cO{\mathcal{O}}
\def\cP{\mathcal{P}}

\def\cU{\mathcal{U}}

\def\ba{\boldsymbol{a}}
\def\bb{\boldsymbol{b}}
\def\bc{\boldsymbol{c}}
\def\bd{\boldsymbol{d}}
\def\be{\boldsymbol{e}}

\def\bm{\boldsymbol{m}}

\def\bu{\boldsymbol{u}}
\def\bv{\boldsymbol{v}}
\def\bw{\boldsymbol{w}}
\def\bx{\boldsymbol{x}}
\def\by{\boldsymbol{y}}
\def\bz{\boldsymbol{z}}
\def\bzero{\boldsymbol{0}}
\def\bone{\boldsymbol{1}}
\def\balpha{\boldsymbol{\alpha}}
\def\bgamma{\boldsymbol{\gamma}}

\def\bbeta{\boldsymbol{\beta}}

\def\bSigma{\boldsymbol{\Sigma}}

\def\rank{\mathrm{rank}}

\def\argmin{\mathrm{argmin}}

\def\supp{\mathrm{supp}}
\def\diag{\mathrm{diag}}

\DeclareMathOperator{\Tr}{\text{Tr}}

\DeclareMathOperator{\OT}{\mathsf{OT}}
\DeclareMathOperator{\SOT}{\mathsf{Sch-OT}}

\newtheorem{theorem}{Theorem}

\newtheorem{proposition}[theorem]{Proposition}

\newtheorem{assumption}[theorem]{Assumption}

\title{Simplifying Optimal Transport through Schatten-$p$ Regularization}
\author{Tyler Maunu \\ Department of Mathematics, Brandeis University \\ \href{maunu@brandeis.edu}{maunu@brandeis.edu}}
\date{\today}

\begin{document}

\maketitle

\begin{abstract}
We propose a new general framework for recovering low-rank structure in optimal transport using Schatten-$p$ norm regularization. Our approach extends existing methods that promote sparse and interpretable transport maps or plans, while providing a unified and principled family of convex programs that encourage low-dimensional structure. The convexity of our formulation enables direct theoretical analysis: we derive optimality conditions and prove recovery guarantees for low-rank couplings and barycentric maps in simplified settings. To efficiently solve the proposed program, we develop a mirror descent algorithm with convergence guarantees for $p \geq 1$. Experiments on synthetic and real data demonstrate the method’s efficiency, scalability, and ability to recover low-rank transport structures.
\end{abstract}

\section{Introduction}

Optimal transport (OT) has emerged as a fundamental computational tool across many areas, including machine learning, computer vision, statistics, and biology \citep{arjovsky2017wasserstein,peyre2019computational,schiebinger2019optimal,bonneel2023survey}. It provides a principled framework for comparing probability distributions, and it has a rich mathematical history \citep{villani2008optimal}. While the combination of practical utility and deep mathematical theory has led to the broad adoption of OT ideas in mathematics, science, and engineering, finding ways to \emph{scale} OT solutions and make them \emph{interpretable} remains a fundamental research question \citep{cuturi2023monge,khamis2024scalable}. In particular, OT typically suffers from the curse of dimensionality \citep{chewi2025statistical}, and regularized estimators may lack sparsity \citep{genevay2019sample}.

A long line of work has focused on making OT scalable and interpretable through \emph{regularization}. The most classical of these is entropic regularization, which yields a strictly convex program that can be solved via Sinkhorn scaling \citep{sinkhorn1967diagonal,cuturi2013sinkhorn}. More recent work has sought to increase efficiency and interpretability through quadratic regularization \citep{blondel2018smooth,lorenz2021quadratically}, as well as low-rank factorizations \citep{forrow2019statistical,scetbon2021low}. These methods show promise in biological applications, particularly in single-cell RNA sequencing analysis \citep{klein2025mapping}.

Another closely related set of recent works attempts to include sparsity in the OT map using \emph{elastic costs} \cite{cuturi2023monge,klein2024learning,chen2025displacement}. In these works, using different cost modifications can be shown to encourage sparse or low-rank transport displacements. This leads to OT maps with simple, interpretable structures. 

Except for entropic regularization, our work simultaneously generalizes all of the aforementioned methods in a unified framework. We believe that this unified picture can lead to more principled development of tailored regularization. Furthermore, the theory of OT has not yet fully leveraged the extensive literature on regularization for scaling and interpretability present in other fields, such as \emph{compressed sensing}. In compressed sensing, the use of $\ell_1$ or nuclear-norm penalties as proxies for rank minimization has yielded provably efficient algorithms \citep{eldar2012compressed,wright2022high}. 
Our general formulation marries ideas from OT and compressed sensing, providing a bridge that we expect to be fruitful for developing sparse and low-rank optimal-transport models moving forward.

\subsection{Contributions}

In this work, we present Schatten-$p$ regularized OT, which we call \emph{Schatten OT}. This novel formulation is both general and amenable to direct theoretical analysis. We summarize the main contributions of our work:
\begin{itemize}
    \item We demonstrate how the Schatten-OT program simultaneously generalizes a large portion of prior work on low-rank and sparse methods in OT, while also yielding new regularized formulations.
    \item We propose a general mirror-descent framework that efficiently solves the Schatten OT scale.
    \item For $p\geq 1$, the resulting optimization problem is convex, allowing convergence guarantees for mirror descent and analysis of low-rank couplings, low-rank transport displacements, and low-rank covariance structures. 
    \item Experiments on synthetic and real data demonstrate the flexibility and effectiveness of  Schatten OT.
\end{itemize}

\subsection{Related Work}

Regularized variants of OT have become increasingly important in current applied and theoretical research. The story of regularized OT begins with entropic regularization \citep{cuturi2013sinkhorn}, which has roots in \cite{schrodinger1932theorie}. More recent regularizations include quadratic and sparse regularization \citep{blondel2018smooth,lorenz2021quadratically,gonzalez2024sparsity}, which seek to encourage sparse structures in the transport plan. 

Other work has studied low-rank factorizations in couplings to scale OT. \cite{forrow2019statistical} define a notion of factored couplings.
\cite{scetbon2021low} use this notion of low-rank factorization of the coupling to develop an efficient Sinkhorn algorithm for factored couplings. Later, \cite{lin2021making} use multiple couplings to move through anchor points.
\cite{halmos2024low} propose a new algorithm to optimize over the LC factorization, and
\cite{halmos2025hierarchical} use hierarchical low-rank structures.

Another line of recent work has studied the regularization of displacements. \citet{cuturi2023monge} introduce the notion of elastic OT costs and show how to construct maps with sparse or low-rank structure. Later, \citet{klein2024learning} introduce learnable parameters into these costs, enabling greater flexibility in selecting the regularizer. \citet{chen2025displacement} use neural networks to learn maps in these settings.

We note that the incorporation of low-dimensional structure in OT displacements dates back to earlier subspace-robust notions of OT. \citet{paty2019subspace} compute Wasserstein distances over worst-case subspaces in the ambient space. These methods have some practical statistical advantages \citep{niles2022estimation}.
 
We can broadly think of regularizing OT as encoding bias in the transport plan. However, there are many other ways the OT problem can be biased. For example, some works seek to encode biases by optimizing the ground cost used within OT. \citet{alvarez2019towards} learn an OT with invariances using an alternating minimization procedure, and focus on optimization over Schatten-$p$ balls.
\citet{sebbouh2024structured} learn a matrix $\bM$ that defines an inner product cost between measures on different spaces.
\citet{jin2021two} match distributions in different spaces using separate linear transformations. We note that these works implicitly regularize transport, as in subspace-robust OT.

In seeking a principled way to regularize and scale OT, we draw connections with \emph{compressed sensing}. Compressed sensing focuses on recovering sparse structures from data. Original foundational works concentrate on recovering sparse vectors using $\ell_1$ regularization \citep{donoho2006compressed,candes2006stable}. These ideas were later extended to low-rank matrices \citep{fazel2008compressed}, which used nuclear norm regularization. The use of more general Schatten-$p$ norms followed this \citep{nie2012low}. The extension of these regularizations to other settings has been fruitful. For example, \cite{scarvelis2024nuclear} use it in the context of deep learning.

The ideas of compressed sensing are seeing a resurgence in the age of modern machine learning and AI. Sparse autoencoders have become a primary tool for practitioners studying mechanistic interpretability \citep{huben2024sparse}. Sparse coding and rate reduction form a recent framework for training deep models to develop ``white-box" methods \citep{yu2020learning,yu2023white}. Compression as a general technique is effective at demonstrating intelligence in simple puzzles \citep{liao2025arcagiwithoutpretraining}. These examples show the importance and practicality of developing theoretically principled compression techniques for machine learning and AI problems. 

\subsection{Notation}

Bold lowercase letters are vectors and bold uppercase letters are matrices.
We denote the set of integers $[n] :=\{1, \dots, n\}$. For vectors, $\|\cdot\|$ is the standard $\ell_2$ (Euclidean) norm. For matrices,  $\|\cdot\|_{S_p}$ is the Schatten-$p$ norm, i.e., the $\ell_p$ norm of the vector of singular values, and $\|\cdot\|_{S_2} = \|\cdot\|_{F}$ is the Frobenius norm. The set of probability measures over $\R^d$ with finite $p$th moment is $\cP_p(\R^d)$, and the subset of absolutely continuous measures is $\cP_{p,ac}(\R^d)$. The indicator function is $\mathbbm{1}$.

\subsection{Organization}

First, in Section \ref{sec:background}, we give the necessary background and outline our optimization program. Then, in Section \ref{sec:algo}, we provide our algorithmic framework for solving the Schatten OT problem.
 After this, Section \ref{sec:theory} gives theoretical results about the structure of Schatten OT couplings. Finally, Section \ref{sec:experiments} presents experiments on synthetic and real data, highlighting the flexibility and advantages of our framework.

\section{Background and Method}
\label{sec:background}

In this section, we first discuss background ideas in OT and compressed sensing, and then the Schatten OT method. We begin in Section \ref{subsec:otreg} by describing discrete OT and its common regularizations. Then, Section \ref{subsec:schatten} discusses background on Schatten-$p$ regularization in compressed sensing.
After this, we define our Schatten-$p$ norm regularized OT, Schatten OT, in Section \ref{subsec:schatten_ot}.

\subsection{OT and Regularization}
\label{subsec:otreg}

For simplicity of presentation, we focus on the discrete case; in Appendix \ref{app:cts_extension}, we show how these ideas extend to the continuous setting. Consider two discrete measures $\mu = \sum_{i=1}^m a_i \delta_{\bx_i}$ and $\nu = \sum_{j=1}^n b_j \delta_{\by_j}$, where $a_i, b_j \geq 0$ and $\sum_i a_i = \sum_j b_j = 1$.  We let $\bX \in \R^{d \times m}$ and $\bY \in \R^{d \times n}$ be matrices with the support points as columns. Without loss of generality, assume $n \geq m$. The transportation polytope $\cU(\ba, \bb)$ is the set of $m \times n$ nonnegative matrices whose rows sum to $\ba = [a_1, \dots, a_m]^\top$ and columns sum to $\bb = [b_1, \dots, b_n]^\top$. We also refer to these matrices as couplings between $\mu$ and $\nu$. In the transport problem, we are thinking of transporting $\mu$ to $\nu$. Therefore, we refer to $\mu$ as the \emph{source distribution} and $\nu$ as the \emph{target distribution}.

We assume an $m \times n$ matrix of costs $\bC$, where $\bC_{ij} = c(\bx_i, \by_j)$, for some function $c: \R^d \times \R^d \to [0, \infty)$. The function $c$ is typically called the \emph{ground cost}. As a concrete example, we can use the $p$th power of the Euclidean distance,
\begin{equation}
    \bC_{ij} = \|\bx_i - \by_j\|^p.
\end{equation}

OT seeks a minimum-cost coupling between the measures $\mu$ and $\nu$. It is formulated as a linear program over the transportation polytope,
\begin{equation}\label{eq:otcost}
    \OT(\mu, \nu) = \min_{\bP \in \cU(\ba, \bb)} \langle \bP, \bC \rangle.
\end{equation}
When the cost corresponds to a power of a metric on some underlying space, the resulting OT cost can be used to define a metric on $\cP_p(\R^d)$. For the rest of this paper, we will assume that $c(\bx_i, \by_j) = \|\bx_i - \by_j\|^2$. However, we note that our regularization can be applied to OT with any ground cost.

For a review of computational methods related to this linear program, see \cite{peyre2019computational}.
While there are many deep and interesting results related to OT, in practice, the direct use of OT in the form presented can suffer. In particular, OT suffers the curse of dimensionality; worst-case statistical rates of estimation for the $p$-Wasserstein distance are $O(n^{-1/d})$, assuming that $\bx_i$ and $\by_j$ are i.i.d. samples from some population measures. On the computational side, for large $n \asymp m$, the linear program incurs computational cost $O(n^3)$, and we must store an $O(n^2)$ variable in memory. To combat these issues, various regularizers have been considered, as we mentioned in the introduction. These regularized OT variants solve
\[
    \min_{\bP \in \cU(\ba, \bb)} \langle \bC, \bP \rangle + \lambda R(\bP),
\]
where $R: \R^{m \times n} \to \R$ is the regularization function and $\lambda$ is a tunable parameter. An example is the entropy function $R(\bP) = \sum_{ij} \bP_{ij} (\log(\bP_{ij}) - 1)$ \citep{cuturi2013sinkhorn}.

\subsection{Schatten-$p$ Regularization in Compressed Sensing}
\label{subsec:schatten}

Schatten-$p$ regularization in compressed sensing served as a way to generalize $\ell_p$ regularization for sparse vector recovery. In particular, using Schatten-$p$ regularization is strictly more general than $\ell_p$ regularization because we can encode vectors as diagonal matrices, in which case $\|\bx\|_{p} = \|\diag(\bx)\|_{S_p}$.

Perhaps the most popular regularization is  Schatten-1 (nuclear norm) regularization. This is typically used to relax the rank of a matrix, and in a variety of settings, nuclear norm minimization has been shown to recover low-rank matrices \citep{recht2010guaranteed,candes2010power,candes2011robust}. 

These methods have been applied in a variety of settings. Some applications of these methods have included matrix completion and recommender systems \citep{nie2012low}, multitask learning \citep{zhang2018learning}, high-dimensional covariance estimation \citep{gavish2017optimal}, and image processing \citep{xie2016weighted}.

Optimization with nuclear norms, or more generally Schatten-$p$ norms, involves a variety of algorithms. For example, nuclear norm minimization can involve saddle point or proximal algorithms \cite{nesterov2013first}, the latter of which involves singular value thresholding \citep{cai2010singular}. Other algorithmic paradigms include primal-dual methods \citep{chambolle2011first} or ADMM \citep{yuan2009sparse}. To solve the more general case of Schatten-$p$ regularized problems, for $0 < p \leq 2$, one typically resorts to Lagrangian-style methods \citep{nie2012low} or iteratively reweighted nuclear norm-style methods \citep{lu2015nonconvex}. Another popular approach to nuclear norm minimization problems involves factor splitting to avoid SVD computations \citep{srebro2004maximum,fan2019factor}.

\subsection{Discrete Schatten-$p$ Regularized OT}
\label{subsec:schatten_ot}

We now come to the main innovation of our work. We study a new variant of regularized OT problems using Schatten-$p$ norms. We define the Schatten OT problem as
\begin{equation}\label{eq:schatten_ot}
   \SOT(\mu, \nu; \{(\lambda_i,p_i,q_i, \cA_i)\}) := \min_{\bP \in \cU(\ba, \bb)} \langle \bC, \bP \rangle + \sum_i \lambda_i \|\cA_i(\bP)\|_{S_{p_i}}^{q_i}.
\end{equation}
The idea of this program is to regularize toward simpler couplings with respect to the maps $\cA_i$.
Notice that the Schatten OT problem relies on three sets of parameters: the regularization strengths $\lambda_i \geq 0$, the Schatten powers and exponents $p_i ,q_i > 0$, and maps $\cA_i: \cU(\ba, \bb) \to \R^{k_i \times l_i}$. Provided that $p_i,q_i \geq 1$ and $\cA_i$ are affine, it is easily seen that Schatten OT is a convex program. With only one regularization term, this simplifies to $\SOT(\mu, \nu; (\lambda, p, q, \cA)) = \min_{\bP \in \cU(\ba, \bb)} \langle \bC, \bP \rangle + \lambda \|\cA(\bP)\|_{S_p}^q$.


We believe that convexity and generality are primary benefits of Schatten OT. It is general enough to cover many existing regularizations in the literature, as we will demonstrate shortly. The convexity of the problem leads to solutions that are easy to characterize, as we will show in our theory section. It also enables efficient solvers using convex optimization techniques.

While many past regularizations for OT fall into this framework, as we will demonstrate below, some do not. In particular, we cannot recover entropic regularization from \eqref{eq:schatten_ot}. While entropic regularization cannot be realized by a Schatten norm of an affine function, one could add an entropic penalty to Schatten OT. This may be convenient from an algorithmic standpoint, but we leave the study of this additional regularization to future work. In Appendix \ref{app:cts_extension}, we illustrate how to extend these ideas to the continuous setting.

\paragraph{Low-rank and Sparse Couplings:} As a first example, consider the affine map $\cA(\bP) = \bP$. Then, depending on the choice of $p$, Schatten OT encourages low-rank or sparse couplings. In particular, choosing $q=p\leq 1$ encourages low-rank solutions. This yields a principled, optimization-based analog to the low-rank factorization pursued by works such as \cite{forrow2019statistical,scetbon2021low}. On the other hand, $q=p=2$ corresponds to the case of quadratically regularized OT \citep{blondel2018smooth,lorenz2021quadratically}, since the Schatten-2 norm is just the Frobenius norm. This tends to encourage sparse solutions \citep{gonzalez2024sparsity}. Group sparsity \citep{blondel2018smooth} can be achieved through proper choice of the affine maps in \eqref{eq:schatten_ot} and setting $p_i=2$, $q_i=1$.

\paragraph{Elastic costs:} We can also recover some of the elastic cost regularizations of \cite{cuturi2023monge, klein2024learning,chen2025displacement}. In particular, we can take $q=p=1$ and let $\cA(\bP)$ be the affine map 
\[
    \bP \mapsto \diag((\bP_{ij}(\bx_i - \by_j))_{i=1, j=1}^{m, n}.
\]
Then, the Schatten OT penalty corresponds to the $\ell_1$ elastic cost. Group-sparse elastic costs can be recovered from sums of Schatten regularizations. We can also recover the subspace elastic costs of \cite{klein2024learning} by taking $q=p=2$ and the affine map
\[
    \bP \mapsto \diag((\bQ_{L}\bP_{ij}(\bx_i - \by_j))_{i=1, j=1}^{n, m},
\]
where $\bQ_{L}$ is the projection onto the orthogonal complement of $L$. Note that, analogously to \cite{klein2024learning}, one could include an additional minimization over $L$ in the Schatten OT formulation. This then defines a family of learnable Schatten OT problems. We discuss this possibility further in the appendix. 

\paragraph{Barycentric projection maps and displacements:} The formulation can be used to penalize map estimators directly. In particular, given a transport plan $\bP$, one can estimate a transport map using the \emph{barycentric projection}
\[
    T_{\bP}(\bx_i) = \frac{1}{a_i} \sum_{j=1}^m \bP_{ij} \by_j = \frac{1}{\ba_i}(\bY \bP^\top)_{:,i}.
\]
Notice that this map is linear in $\bP$. Thus, we can penalize the barycentric projection map in our program by letting $\cA(\bP) = T_{\bP}(\bX) \diag(\bA)^{-1/2}$, where we define $\bA = \diag(\ba)$. As a further example, we could encourage displacements to be low-rank rather than the map itself. We call $T_{\bP}(\bx_i) - \bx_i$ the \emph{barycentric displacement}. Then, to encourage these to be simple, we could use $\cA(\bP) =  \bY \bP^\top \bA^{-1/2}   - \bX \bA^{1/2}$. In both cases, the additional scaling of $\bA^{1/2}$ allows the population limit of the program to be well defined.


\paragraph{Covariance regularization:} All of the maps discussed so far include zeroth and first moments of the support points $(\bx_i, \by_j)$. However, our formulation is flexible enough to include higher moments of our data distribution. For example, we can take $\cA$ to be an affine function of the covariance induced by $\bP$, \[
    \bSigma_{\bP} = \sum_{ij} \bP_{ij} \begin{pmatrix}
        \bx_i \\ \by_j
    \end{pmatrix}\begin{pmatrix}
        \bx_i \\ \by_j
    \end{pmatrix}^\top,
\]
which is linear in $\bP$. 

We could penalize the Schatten-1 norm of the cross-covariance $\sum_{ij} \bP_{ij} \bx_i \by_j^\top$, which is an affine function of $\bSigma_{\bP}$. If the vectors $\bx_i$ and $\by_j$ are whitened (i.e., they each have identity covariance), then the singular values of this matrix correspond to the canonical correlations. Minimizing the Schatten-1 norm in this case corresponds to minimizing the sum of canonical correlations, which seeks to increase independence between $X$ and $Y$. Note that if we take the $\cA(\bP) = \sum_{ij} \bP_{ij} (\bx_i - \by_j) (\bx_i - \by_j)^\top$ and $p=1$, then the regularization is just the quadratic OT cost, $\|\sum_{ij} \bP_{ij} (\bx_i - \by_j) (\bx_i - \by_j)^\top\|_{S_1} = \sum_{ij} \bP_{ij} \|\bx_i - \by_j\|^2$. 

These illustrate just a few of the potential choices for adding covariance regularization to OT. In the appendix, we discuss covariance regularization in the context of Schatten OT for Gaussians. An in-depth study of these is left to future work.

\section{A Mirror Descent Algorithm}
\label{sec:algo}

The Schatten-OT program in \eqref{eq:schatten_ot} is convex whenever $p_i,q_i \geq 1$ and $\cA_i$ are affine, but solving it directly with off-the-shelf convex solvers (e.g., CVXPY \citep{diamond2016cvxpy} or interior point methods) is only feasible for small problems, as the transportation polytope $\cU(\ba,\bb)$ involves $\cO(nm)$ variables and constraints. To address large-scale settings, we turn to first-order optimization methods. A particularly effective choice for optimization over the transport polytope is mirror descent with Kullback–Leibler (KL) geometry \cite{kemertas2025efficient}. We use this algorithm for its simplicity and leave the analysis of more general methods, such as primal-dual algorithms or ADMM, to future work. 

Following the approach of \citet{kemertas2025efficient}, we develop mirror descent using the KL geometry on the transport polytope.  
This choice is natural, since using the KL geometry replaces a costly Euclidean projection with efficient Sinkhorn scaling. A few iterations of this method, followed by rounding, are effective at projecting to the polytope \citep{altschuler2017near}.

Assuming that $\cA(\bP) \neq 0$, we can compute a subgradient of the Schatten-$p$ norm term in the Schatten OT problem as
\[
    q\|\cA(\bP)\|_{S_p}^{q-p}\cA^\star \left(\bU \bSigma^{p-1} \bV^\top\right) \in \partial \|\cA(\bP)\|_{S_p}^q,
\]
where $\partial$ denotes the subdifferential. This provides a computable subgradient of $F(\bP)$ at each iteration, provided that we can compute a singular value decomposition. For $p,q > 1$, this is a bona fide gradient.

The mirror descent iteration involves the following steps.
\begin{enumerate}
    \item Form the SVD $\cA(\bP^k) = \bU^k \bSigma^k \bV^{k\top}$ and the subgradient
    \[
        \bG^k = q\|\cA(\bP^k)\|_{S_p}^{q-p}\cA^\star\!\left(\bU^k (\bSigma^k)^{p-1} \bV^{k\top}\right)
    \]
    \item Use multiplicative-weights form
    \[
    \widehat{\bP}_{ij}  \propto  \bP^k_{ij} \exp(-\tau \bG^k_{ij})
    \]
    \item Project back to the transport polytope
    \[
    \bP^{k+1} = \Pi^{\mathrm{KL}}_{\cU(\ba,\bb)}(\widehat{\bP}),
    \]
\end{enumerate}
Here, $\tau^k>0$ is a step size and $\Pi^{\mathrm{KL}}_{\cU(\ba,\bb)}$ is the projection onto the transport polytope with respect to the KL divergence, which can be implemented via Sinkhorn scaling.

By standard mirror descent theory \citep{beck2003mirror,nemirovsky1983problem,bubeck2015convex}, the method achieves an $O(1/\sqrt{T})$ convergence rate for convex objectives ($p,q \geq 1$).  
The KL geometry ensures that nonnegativity is automatically preserved, and averaging can be used to guarantee convergence of the objective values.

The most expensive parts of the iteration are the SVD computation and the Sinkhorn projection. It is possible to use an adaptive low-rank approximation of $\cA(\bP^k)$ throughout the iterations to increase computational efficiency. It would also be interesting to attempt to use sketching methods to approximate low-rank solutions to this problem \citep{yurtsever2021scalable}.

The choice of step size is essential. In general, since the problem is convex but not smooth in general, one could take $\tau^k \propto 1/\sqrt{k}$, which yields the $O(1/\sqrt{T})$ convergence rate in objective value.
In our experiments, we can observe faster convergence in specific settings. For example, when $p=q=1$ and $\cA$ has simple structure, for instance when $\cA(\bP)=\bP$ (low-rank couplings) or $\cA(\bP)= \bY \bP^\top \bA^{-1}$ (low-rank barycentric maps), mirror descent can obtain faster convergence with a geometrically diminishing step size, the same schedule used in past work with sharp minima \citep{davis2018subgradient,maunu2019well}.  

\section{Theory}
\label{sec:theory}

In this section, we present our main theoretical results on the Schatten OT program.
First, Section \ref{subsec:theory_struct} uses convex optimization theory to outline the structure of solutions to the Schatten OT problem. After this, Section \ref{subsec:theory_recovery} uses this structure to prove two theorems that demonstrate Schatten OT's ability to recover low-rank couplings and barycentric displacements. Finally, in Section \ref{subsec:theory_discussion}, we finish with a discussion of our theoretical results.

\subsection{General Structural Theorems}
\label{subsec:theory_struct}

Let $\bP^\star$ be an optimal solution of the optimization problem \eqref{eq:schatten_ot} with $p,q\geq1$.
Since this is a constrained convex optimization problem in $\bP$, we can appeal to standard theory. 
The solution is characterized by the KKT conditions, which state that there exists $\bG^\star \in \partial \|\cA(\bP^\star)\|_{S_p}^q$ such that
\begin{align*}
    &\bC + \lambda  \bG^{\star } + \bone_n \bu^\top + \bv \bone_m^\top = \bzero, \\
    & \bP^\star \geq 0, \quad \bP^\star \bone = \ba, \quad \bP^{\star\top} \bone = \bb.
\end{align*}
Comparing these conditions to the optimality conditions for standard OT, we notice that the only difference is the inclusion of the $\lambda \bG^\star$ in the first-order stationarity condition. Therefore, the  optimality conditions for Schatten OT are precisely those for an OT problem with the \emph{tilted cost} 
\begin{equation}\label{eq:S-def}
\bS(\lambda,\bG^\star) := \bC + \lambda  \bG^{\star }  \in \R^{n\times m}.
\end{equation}
where $\bG^\star$ is some subgradient of $\|\cA(\cdot)\|_{S_p}^q$ at $\bP^\star$. The obstacle to our directly applying this result is that we do not know $\bG^\star$, since that would require knowing $\bP^\star$.

We can state this characterization as the following proposition.
\begin{proposition}
    The coupling $\bP^\star$ is optimal for \eqref{eq:schatten_ot} if and only if there exists a subgradient $\bG^\star$ of $\|\cA(\bP^\star)\|_{S_p}^q$ such that
    \[
        \bP^\star \in \argmin_{\bP \in \cU(\ba, \bb)} \langle \bS(\lambda, \bG^\star),\bP\rangle.
    \]
\end{proposition}

While we cannot apply this to the direct computation of $\bP^\star$, we can use it to characterize solutions to the Schatten OT problem. In the following section, we develop this idea to prove the recovery of low-rank structure in the Schatten OT problem.

\subsection{Discrete Recovery Theorems}
\label{subsec:theory_recovery}

In this section, we prove low-rank recovery theorems for Schatten OT. While these are restrictive toy examples, they represent the first such exact recovery results in the OT literature. We believe this is a first step towards applying compressed sensing ideas to regularized OT problems. It is an open question for future work to extend these ideas to the recovery of simple couplings in more complex settings.  
We begin in Section \ref{subsubsec:theory_lowrank_coupling} with a recovery result for low-rank couplings. Then, Section \ref{subsubsec:theory_lowrank_displacement} presents a consequence on the recovery of a low-rank set of barycentric displacements.

\subsubsection{Low-Rank Coupling Recovery}
\label{subsubsec:theory_lowrank_coupling}

We now assume that both the source and the target consist of $R$ well-separated clusters, each with the same cardinality. We show that, for a nontrivial interval of regularization strengths $\lambda$, the nuclear-norm penalized OT problem recovers a rank-$R$, block-diagonal coupling that matches each source cluster uniformly to its corresponding target cluster. We assume uniform marginals $a_i = 1/m, b_j=1/n$ for all $i,j$. 

Our first assumption is on the clustered structure of $\mu$ and $\nu$.
\begin{assumption}\label{assump:clustered}
    For two measures $\mu = \sum_{i=1}^m a_i \delta_{\bx_i}$ and $\nu = \sum_{j=1}^n b_j \delta_{\by_j}$, $n=Rg$, $m=Rg$ for integers $R,g \geq 1$. The source indices $[m]$ and target indices $[n]$ are partitioned into clusters $S_1,\dots,S_R$ and $T_1,\dots,T_R$, respectively, where $|S_t|=|T_t|=g$ for all $t$.
\end{assumption}
 Let $B(\bz,\rho)$ denotes the closed Euclidean ball of radius $\rho>0$ around $\bz\in\R^d$. Our goal is to construct a setting where the cluster $S_t$ is uniformly matched to $T_t$. We make the following assumptions about the locations of the source and target points. 
\begin{assumption}\label{assump:separation}
    For two measures $\mu = \sum_{i=1}^m a_i \delta_{\bx_i}$ and $\nu = \sum_{j=1}^n b_j \delta_{\by_j}$,
    \begin{enumerate}
        \item The source points lie in disjoint balls, $\bx_i \in B(\bc_t,\rho)$ for $i\in S_t$, and the target points lie in disjoint balls, $ \by_j\in B(\bd_t,\rho)$ for $j\in T_t$.
        \item Within matched clusters $S_t$ and $T_t$, $\|\bx_i - \by_j\| = \|\bx_i - \by_{j'}\|$ for all $i \in S_t$ and $j, j' \in T_t$ for $t=1, \dots, R$.
        \item The minimum inter-cluster distance $ \Gamma := \min_{s\neq t} \|\bc_t-\bd_s\|$ , and the maximum intra-cluster distance as $\gamma := \max_{t} \|\bc_t-\bd_t\|$ satisfy      \begin{equation}\label{eq:sep-margin}
\Gamma\ >\ \gamma + 4\rho>0.
\end{equation}
        
    \end{enumerate}
\end{assumption}

Notice that, under our separation condition \eqref{eq:sep-margin}, the OT coupling actually respects the cluster structure, in the sense that it must match points in $S_t$ to $T_t$. Furthermore, any plan that matches $\bx_i$ to $\by_j$ within clusters (when $i \in S_t$, $j \in T_t$) is optimal. However, these matched clusterings are not low-rank; they are full-rank. On the other hand, as we will show in the following theorem, a low-rank matching can be recovered from Schatten OT.

\begin{theorem}\label{thm:R-vs-R}
Let Assumptions \ref{assump:clustered} and \ref{assump:separation} hold, and let the excess cost for an across-cluster matching be
\[
\Delta_{\min}  := \min_{ \substack{s \neq t \in [R] \\ i \in S_t, j \in T_s, j' \in T_t}}
\Big\{ \|\bx_i-\by_j\|_2^2 - \|\bx_i-\by_{j'}\|_2^2 \Big\}.
\]
Then, for any regularization parameter $\lambda$ satisfying
\begin{equation}\label{eq:lambda-window}
0\ \le\ \lambda\ <\ g\cdot \Delta_{\min},
\end{equation}
the minimizer of $\langle \bC,\bP\rangle+\lambda\|\bP\|_{S_1}$ is a rank $R$ coupling supported blockwise on
$\bigcup_{t=1}^R (S_t\times T_t)$ that is uniform within clusters. 
\end{theorem}

The essential idea of the proof is to ensure that 1) $\bP^\star$ is the unique coupling that respects the cluster structure and also minimizes the nuclear norm, and 2) there is no way to make the nuclear norm term even smaller by using across-cluster matches without incurring more cost.

\subsubsection{Low-Rank Displacement Recovery when $p=1$}
\label{subsubsec:theory_lowrank_displacement}

We now give a concrete example of how to recover a coupling with low-rank displacements. For our affine map, we consider the weighted barycentric displacement matrix $\cA(\bP)$,
\[
\cA(\bP)\ :=\ \big(T_{\bP}(\bX)-\bX\big) \bA^{1/2}\ = \bY \bP^\top \bA^{-1/2} - \bX \bA^{1/2}.
\]
We also again assume that $p=q=1$ in the Schatten OT formulation. For our recovery result, we make the following assumptions.

\begin{assumption}[Symmetric two-target clusters with separation]\label{assump:sym_pairs}
Fix orthonormal vectors $\bu,\bv\in\R^d$ and an integer $R\geq 2$.
Let $0<\mu_1<\dots<\mu_R$ be distinct scalars and put $\bm_t:=\mu_t \bu \in \R^d$ for $t\in[R]$.
Suppose:
\begin{enumerate}
\item The source set $[m]$ is partitioned into nonempty clusters $S_1,\dots,S_R$ and there exists $\rho>0$ such that
$\bx_i\in \bm_t + [-\rho,\rho] \bu$ for all $i\in S_t$.
\item For some $\varepsilon>0$, the target support consists of the $2R$ points
\[
\by_{t,+}=\bm_t+\varepsilon \bv,\ \by_{t,-}=\bm_t-\varepsilon \bv, t\in[R],
\]
with target masses $b_{t,+}=b_{t,-}=\frac{1}{2} \sum_{i\in S_t} a_i$.
\item The minimal separation between clusters is lower bounded
\[
\Lambda\ :=\ \min_{s\neq t} |\mu_t-\mu_s|\ >\ 2\rho.
\]
\end{enumerate}
\end{assumption}

By symmetry, it is easy to see that all couplings that assign the source points in cluster $S_t$ to $\by_{t,+}$ or $\by_{t,-}$ are optimal. In particular, it does not matter how the points are assigned within clusters. For $i\in S_t$ and $s\neq t$, it is also convenient to define the inter-cluster cost gap
\begin{equation}\label{eq:baseline-gap}
\Delta_{i,s}\ :=\ \|\bx_i-\by_{s,\pm}\|_2^2 - \|\bx_i-\by_{t,\pm}\|_2^2
\ =\ (\mu_t-\mu_s)^2 + 2(\mu_t-\mu_s) \xi_i,
\end{equation}
where $\bx_i=\bm_t+\xi_i \bu$ and $|\xi_i| \leq \rho$. 

We now state the main recovery theorem for this setting. It says that, under our assumption, we can exactly recover a coupling with a rank-1 barycentric displacement.

\begin{theorem}\label{thm:rank1-recovery}
Under Assumption~\ref{assump:sym_pairs} and the quadratic cost $\bC_{ij} = \|\bx_i - \by_j\|^2$. Define the explicit threshold
\[
\lambda_{\max}\ :=\ \Lambda-2\rho\ >\ 0.
\]
Then for every $\lambda\in[0,\lambda_{\max})$, the unique minimizer of~\eqref{eq:schatten_ot} matches $\bx_i$, for $i \in S_t$, to $\{ \by_{t,\pm}\}$. Furthermore, it yields a rank-$1$ barycentric map.
\end{theorem}

\subsection{Discussion}
\label{subsec:theory_discussion}

While we demonstrate low-rank recovery only in toy examples here, our methodology highlights the advantages of using convex formulations. In particular, it is easier to verify that the recovered solution is low-rank. In fact, these are the first guarantees of low-rank recovery within an OT problem in the literature. This is compared to nonconvex methods, which currently lack guarantees \citep{forrow2019statistical,klein2024learning}.

In the future, it would be interesting to demonstrate exact recovery in more general settings using Schatten-$p$ regularization when $p < 1$. It would also be interesting to develop more general recovery conditions to move beyond the toy examples considered here.

\section{Experiments}
\label{sec:experiments}

In this section, we give some simulations on real data that demonstrate the advantages of the Schatten OT formulation. First, in Section \ref{subsec:low_rank_recovery}, we demonstrate Schatten OT's ability to recover low-rank couplings and barycentric projection maps. Then, in Section \ref{subsec:convergence_rates}, we examine the convergence rate of mirror descent to solve the Schatten OT problem. Finally, Section \ref{subsec:4i} gives an experiment on real data that demonstrates the ability of Schatten OT to recover simpler couplings with 4i perturbation data.

\subsection{Low-rank Recovery}
\label{subsec:low_rank_recovery}

To first examine properties of the Schatten OT problem, we use CVXPY to solve the convex program exactly.

In our first experiment, we examine the Schatten OT's ability to recover low-rank couplings. To measure the quality of the recovered $\bP^\star$, we use two metrics: effective rank and transport cost. The latter is defined as $\langle \bC, \bP^\star \rangle$. The former is the ratio of the nuclear norm to the operator norm:
\[
    \textsf{Effective  Rank}(\bB) = \frac{\|\bB\|_{S_1}}{\|\bB\|_{S_{\infty}}}.
\]

In our experiments, the support of $\mu$ consists of two clusters centered at $(-2, 2)$ and $(-2, -2)$, and $\nu$ consists of two clusters centered at $(2, 2)$, and $(2,-2)$. The data within each cluster is Gaussian. We sample 10 points from each cluster, so that $n=m=20$. The results are averaged over five randomly generated datasets. 

\begin{figure}[ht]
    \centering
    \includegraphics[width=0.35\linewidth]{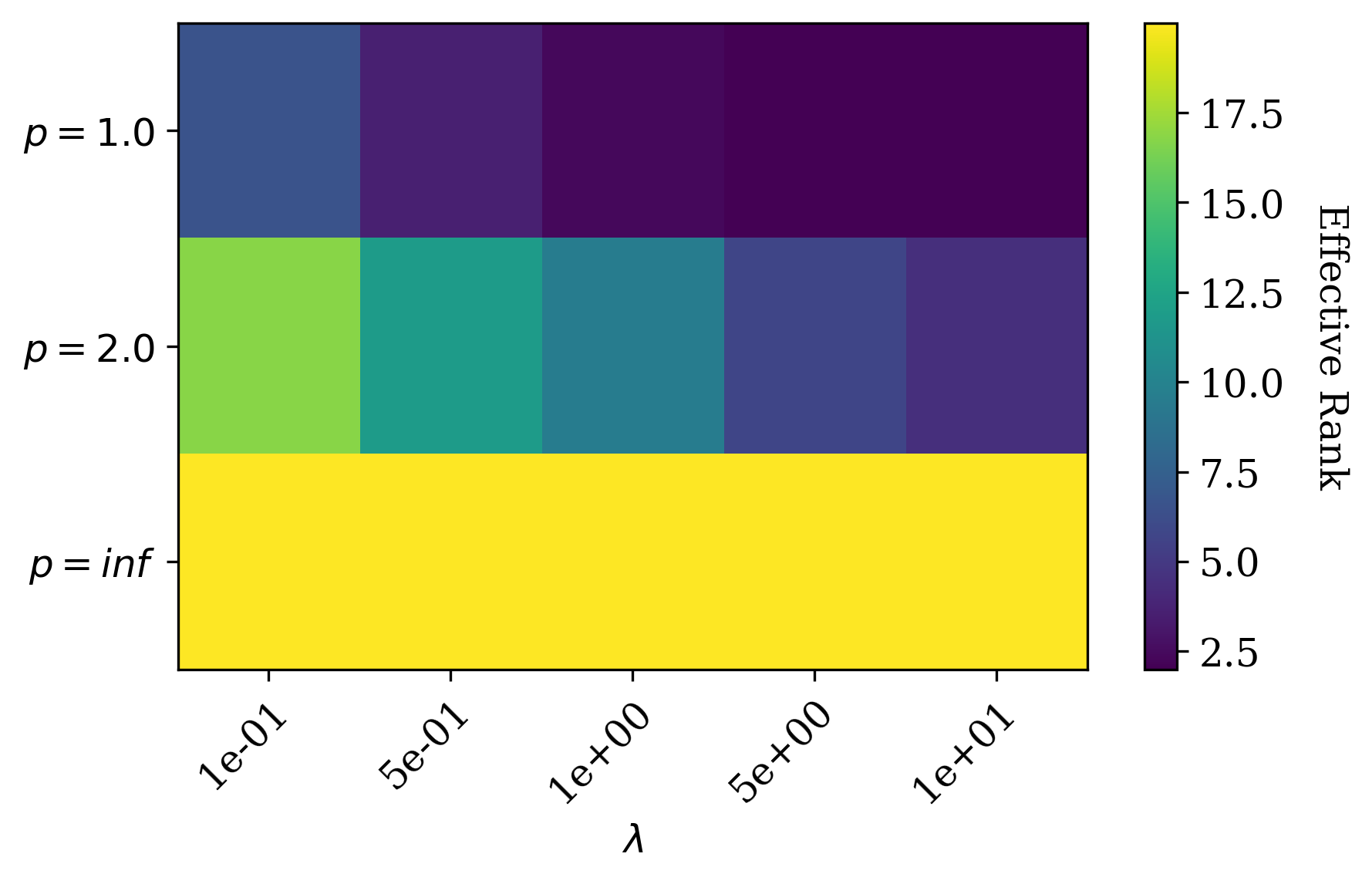}
    \includegraphics[width=0.35\linewidth]{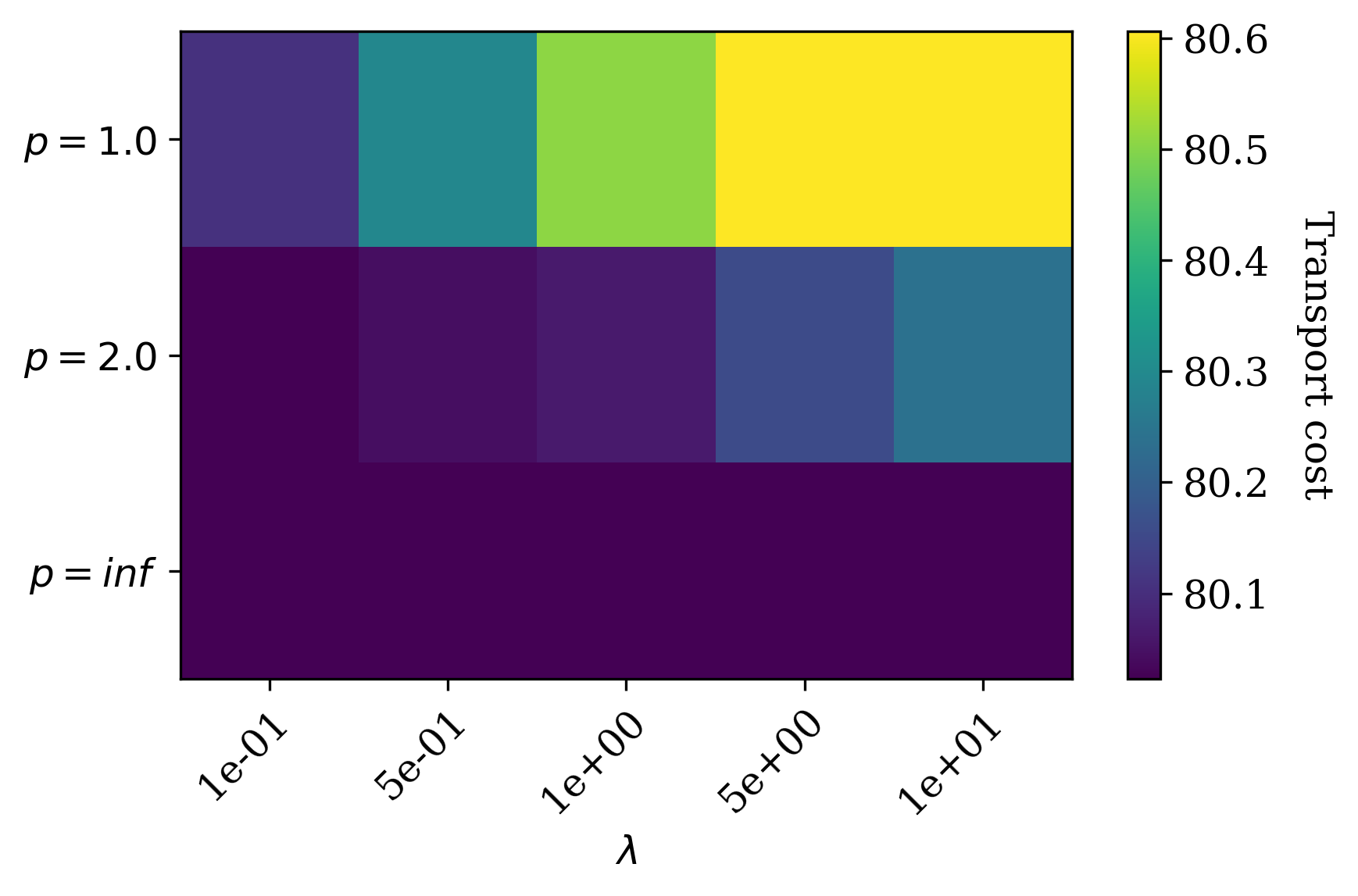}
    \includegraphics[width=0.35\linewidth]{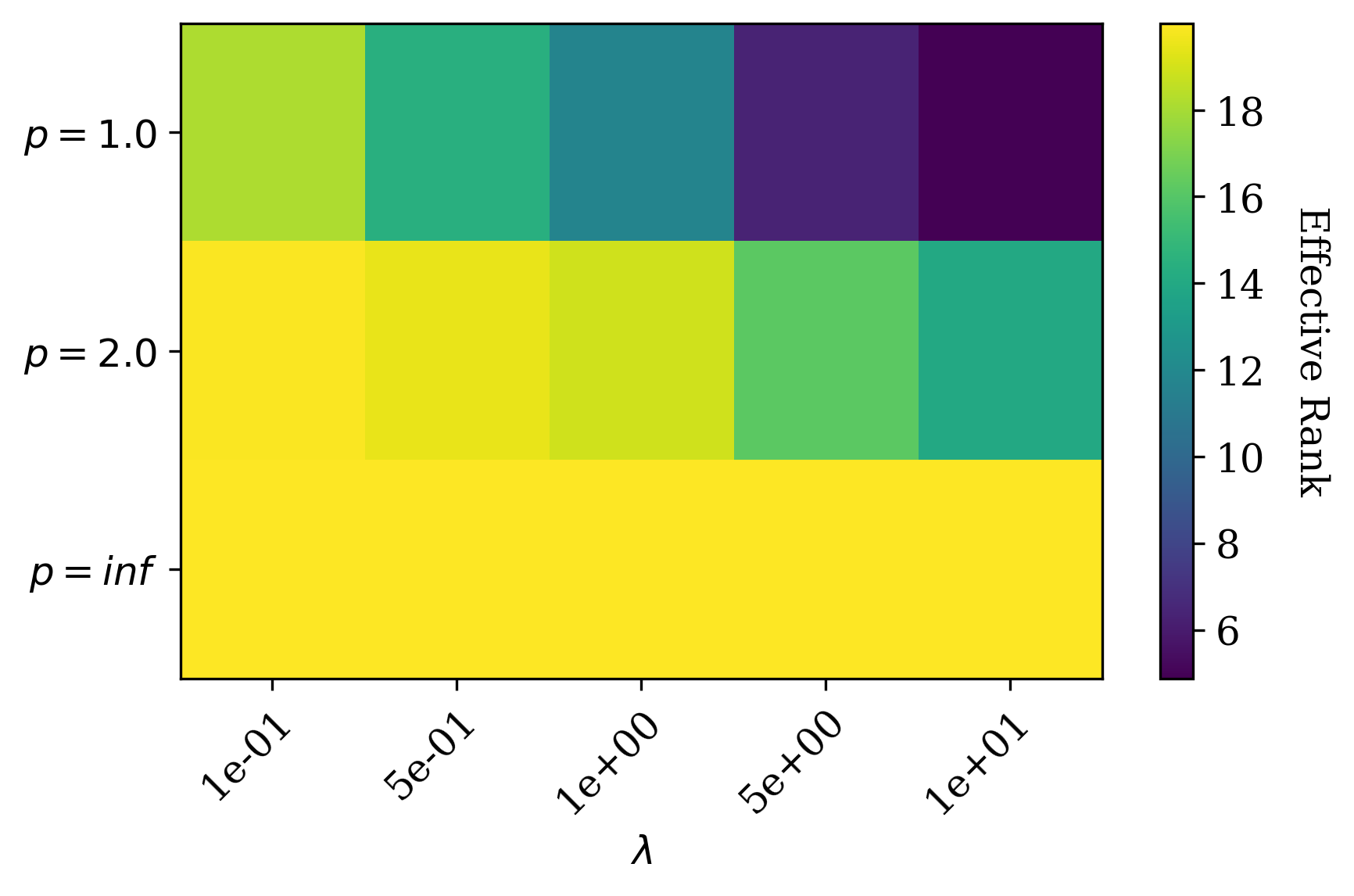}
    \includegraphics[width=0.35\linewidth]{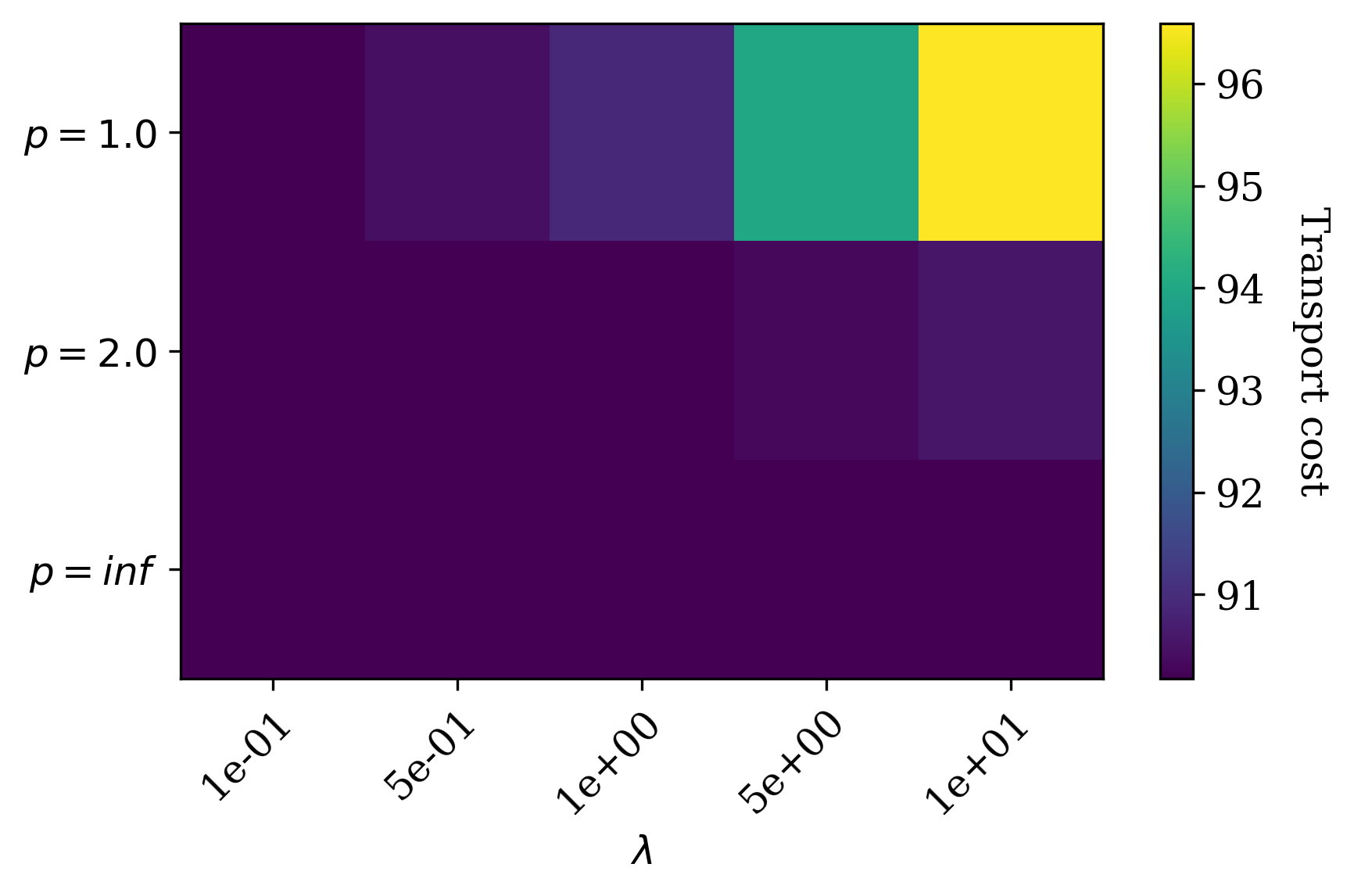}
    \caption{Solution quality of Schatten OT versus regularization parameter for mixture of Gaussian data. Top row: small variance. Bottom row: large variance. On the left, we show the effective rank of the found solution; on the right, we display its transport cost. As we can see, Schatten-1 regularization can greatly simplify the transport plan without substantially increasing transport costs.}
    \label{fig:low_rank_coupling}
\end{figure}

In the first experiment displayed in the top row of Figure \ref{fig:low_rank_coupling}, we use a variance of 0.04 for each Gaussian component. In the left image, we show the effective rank versus regularization strength $\lambda$ for $p=1, 2$, and $\ infty$. On the right, we display the transport cost. As we can see, nuclear norm regularization can significantly reduce the effective rank without substantially increasing the transport cost. Using the Schatten-2 norm can also reduce the effective rank, albeit more gradually. 

We compare this experiment with a slight modification in the bottom row of Figure \ref{fig:low_rank_coupling}. Here, the data model is the same, except now the within-cluster variance is $2$. As we can see, it is more challenging to find a low-rank transport plan, and when one is found, it increases the transport cost more substantially.

\begin{figure}[ht]
    \centering
    \includegraphics[width=0.35\linewidth]{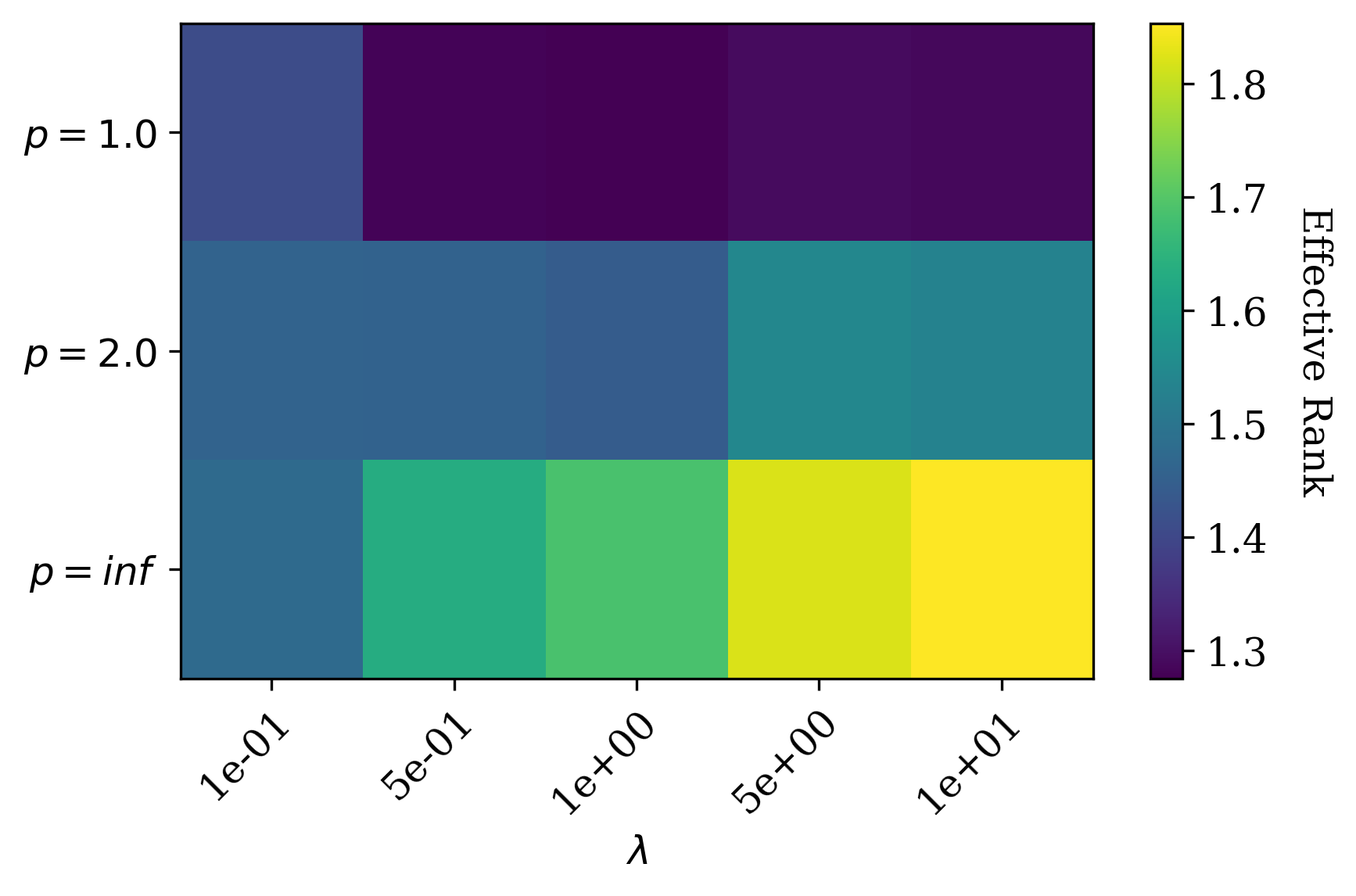}
    \includegraphics[width=0.35\linewidth]{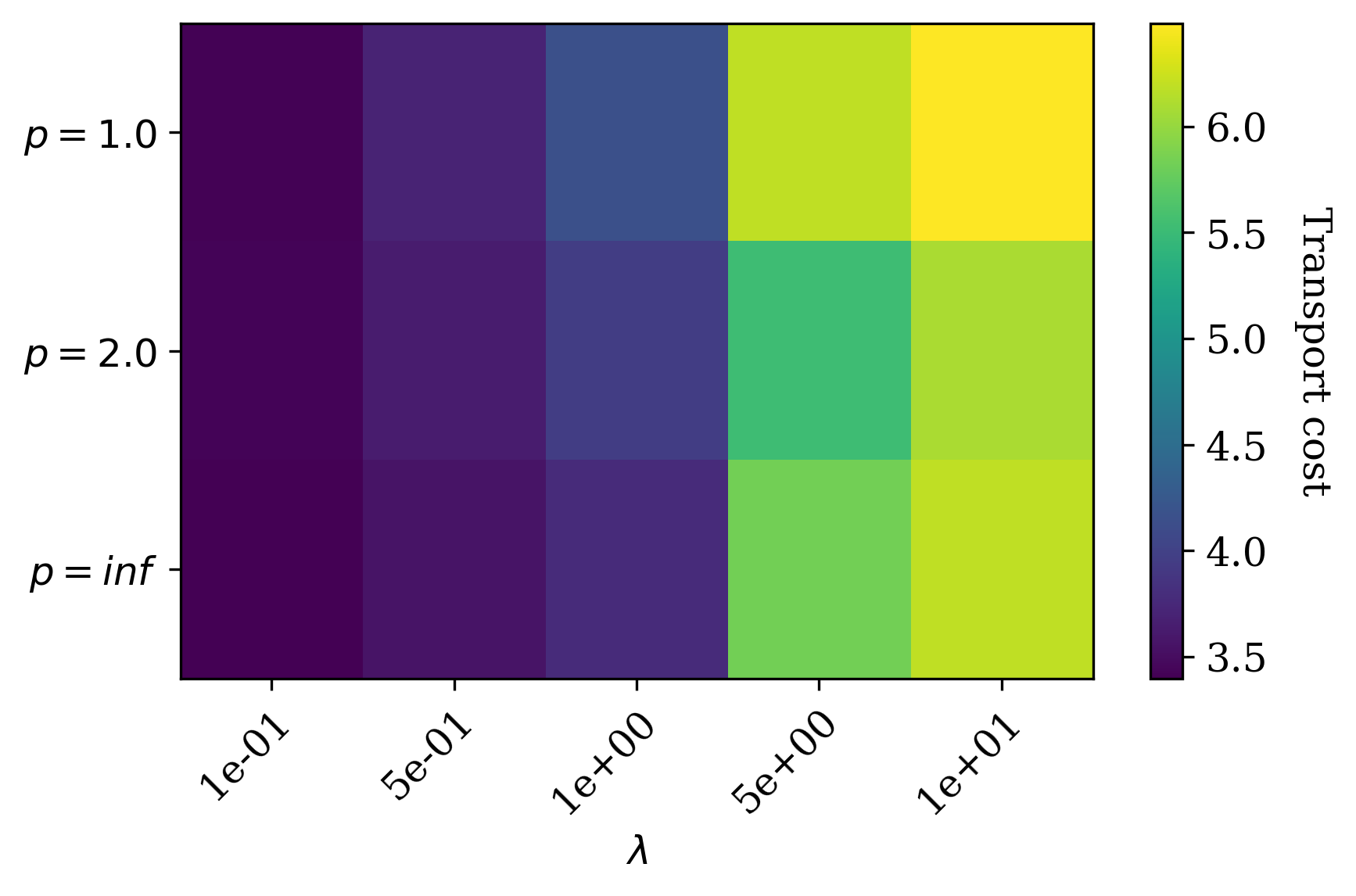}
    \caption{Solution quality of Schatten OT versus regularization parameter for Gaussian data with a low-rank perturbation. In the left display, we show the effective rank of the found barycentric displacements, and in the right display, we show the transport cost of the found coupling. As we can see, Schatten-1 regularization again simplifies the transport plan, but the transport cost increases substantially across the board.}
    \label{fig:low_rank_bary}
\end{figure}

We include one more experiment in which we now wish to recover a low-rank displacement. We assume that the support of $\mu$ is standard Gaussian, and the support of $\nu$ is $\by_i = \bx_i + \xi_i \bu$, where $\bu$ is a random unit vector and $\xi_i$ is standard Gaussian.  We now wish to recover a coupling with a low-rank barycentric map, and do the same experiment with a different affine map of $\bP$. Figure \ref{fig:low_rank_bary} displays the results of this experiment. As we can see, Schatten-1 regularization again recovers a coupling with lower-rank displacements. However, the transport costs increase more substantially across the board with higher regularization strengths.

\subsection{Convergence Rates}
\label{subsec:convergence_rates}

In this section, we examine the convergence rate of the mirror descent algorithm with Schatten-1 regularization in two settings. In the first setting, we show sublinear convergence; in the second, linear convergence.

\begin{figure}[ht]
    \centering
    \includegraphics[width=0.35\linewidth]{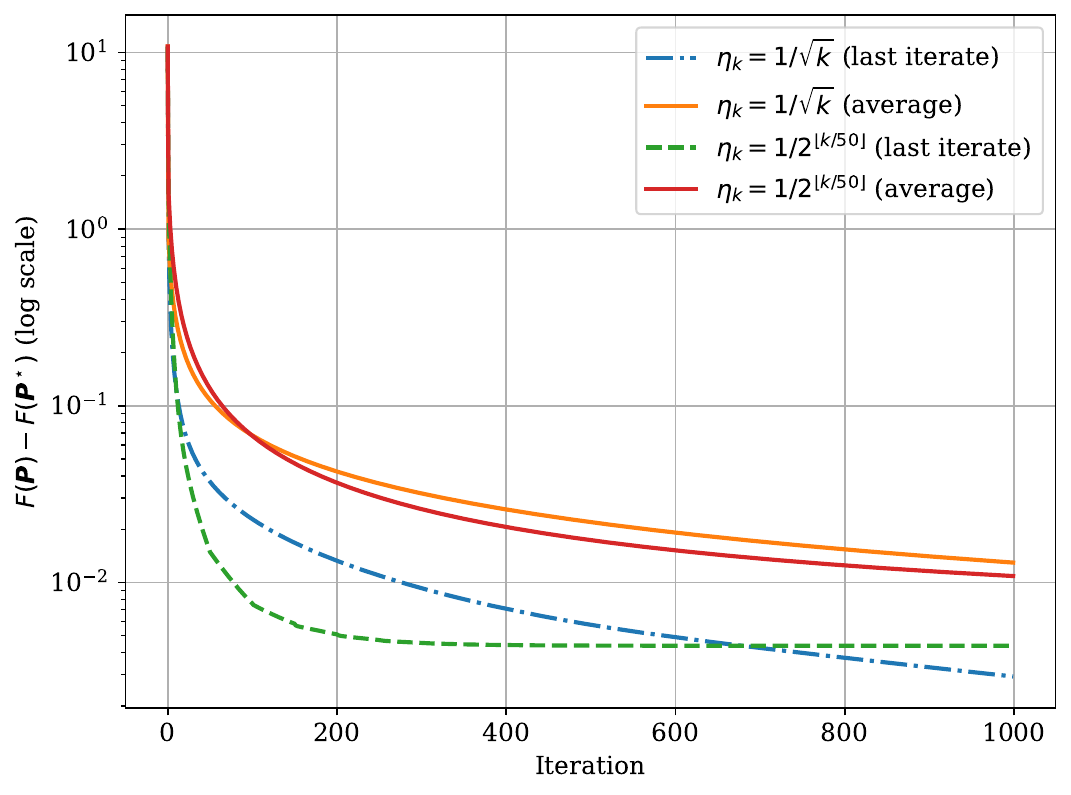}
    \includegraphics[width=0.35\linewidth]{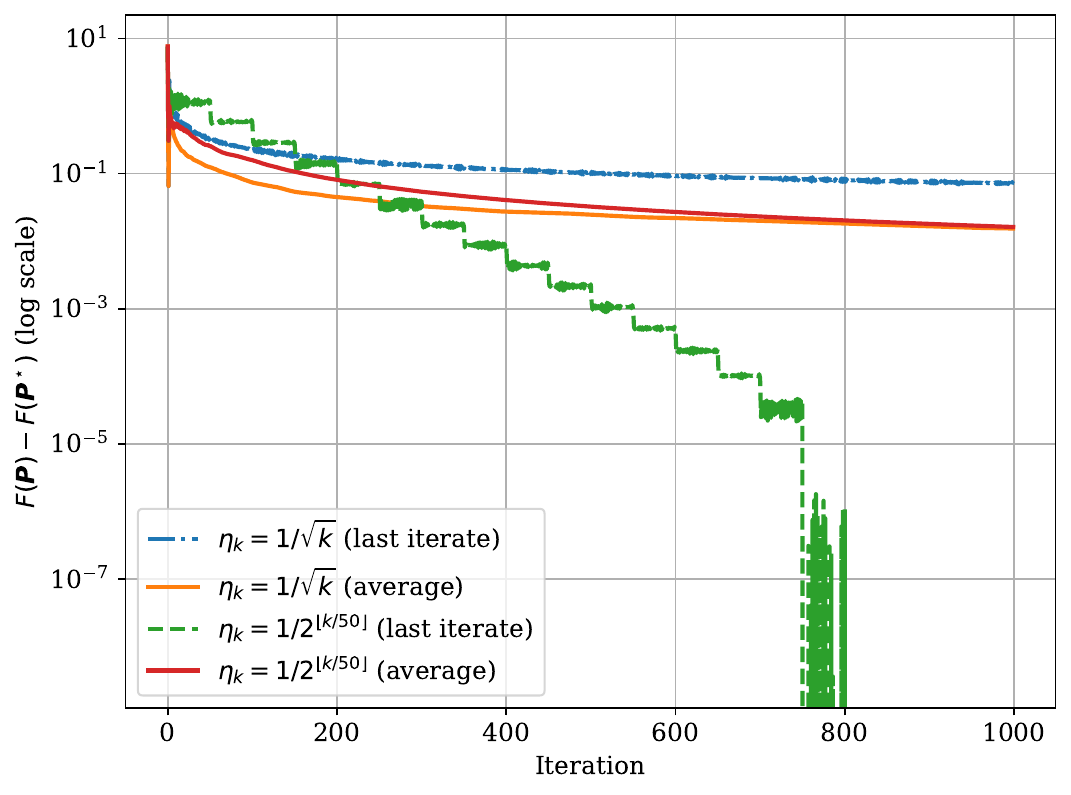}
    \caption{Plot of log excess cost versus iteration for the mirror descent algorithm on the Schatten OT problem. Left: In this experiment, the regularization parameter is small, and the variance of the Gaussian mixture components is large. This shows sublinear convergence of the algorithm, as is expected by the theory. Right: we reduce the variance and increase the regularization parameter, resulting in an optimal low-rank coupling. Here, we see that the geometrically diminishing step size converges linearly.}
    \label{fig:convergence1}
\end{figure}

In the left display of Figure \ref{fig:convergence1}, we use a setting where we do not expect a low-rank coupling to be easy to find. We set $\lambda=.1$, and the data $\mu$ is a mixture of Gaussians with centers at $(-2, \pm 2)$ and $\nu$ is a mixture of Gaussians with centers at $(2, \pm 2)$. The variance is set to be 1, and $n=m=20$. We observe slow sublinear convergence. Note the log scale on the x-axis.

In the right display of Figure \ref{fig:convergence1}, we use the same setup as before, except now we set $\lambda=10$ and let the clusters have variance $0.04$. Now, since the recovered coupling is low-rank, mirror descent with a geometrically diminishing step size converges linearly. This implies that the objective is sufficiently sharp, which can be exploited by this step-size schedule.

\subsection{4i Perturbation Example}
\label{subsec:4i}

In the experiment of \cite{chen2025displacement}, the authors fit a displacement-sparse neural OT to 4i perturbation data. In the experiment, we see that dimensionality is reduced, but the error is higher, and the method has high variance because it requires fitting an input convex neural network (ICNN).

In the following, we show how Schatten OT can reduce the effective rank of couplings and barycentric projection maps. We use two perturbations within the CellOT data of \cite{bunne2023learning}. In particular, we follow \cite{chen2025displacement} and consider learning regularized couplings from the 4i perturbation data. The processed data is publicly available\footnote{\url{https://www.research-collection.ethz.ch/handle/20.500.11850/609681}}. More details on our algorithmic setup for this experiment are given in Appendix \ref{app:exp_details}.

In Figure \ref{fig:4i_lowrank}, we plot the effective rank against $\lambda$ for two different affine maps $\cA(\bP) = \bP$ and $\cA(\bP) = \bY \bP^\top \bA^{-1/2}$. The color indicates the increase in transport cost relative to Sinkhorn with a regularization parameter of 1. We display the result for two different perturbations. For each, we average over five random subsamples of size 1000 from the control and perturbation distributions. 

On the top row, we display the result for low-rank coupling recovery. As we can see, we can drastically reduce the complexity of transport plans without paying much more in transport costs. Next, in the bottom row of Figure \ref{fig:4i_lowrank}, we repeat the previous experiment, focusing on recovering a low-rank barycentric projection map. We see again that Schatten-1 regularization can reduce the effective rank, although now the transport cost increases more substantially.
\begin{figure}[ht]
    \centering
    \includegraphics[width=0.35\linewidth]{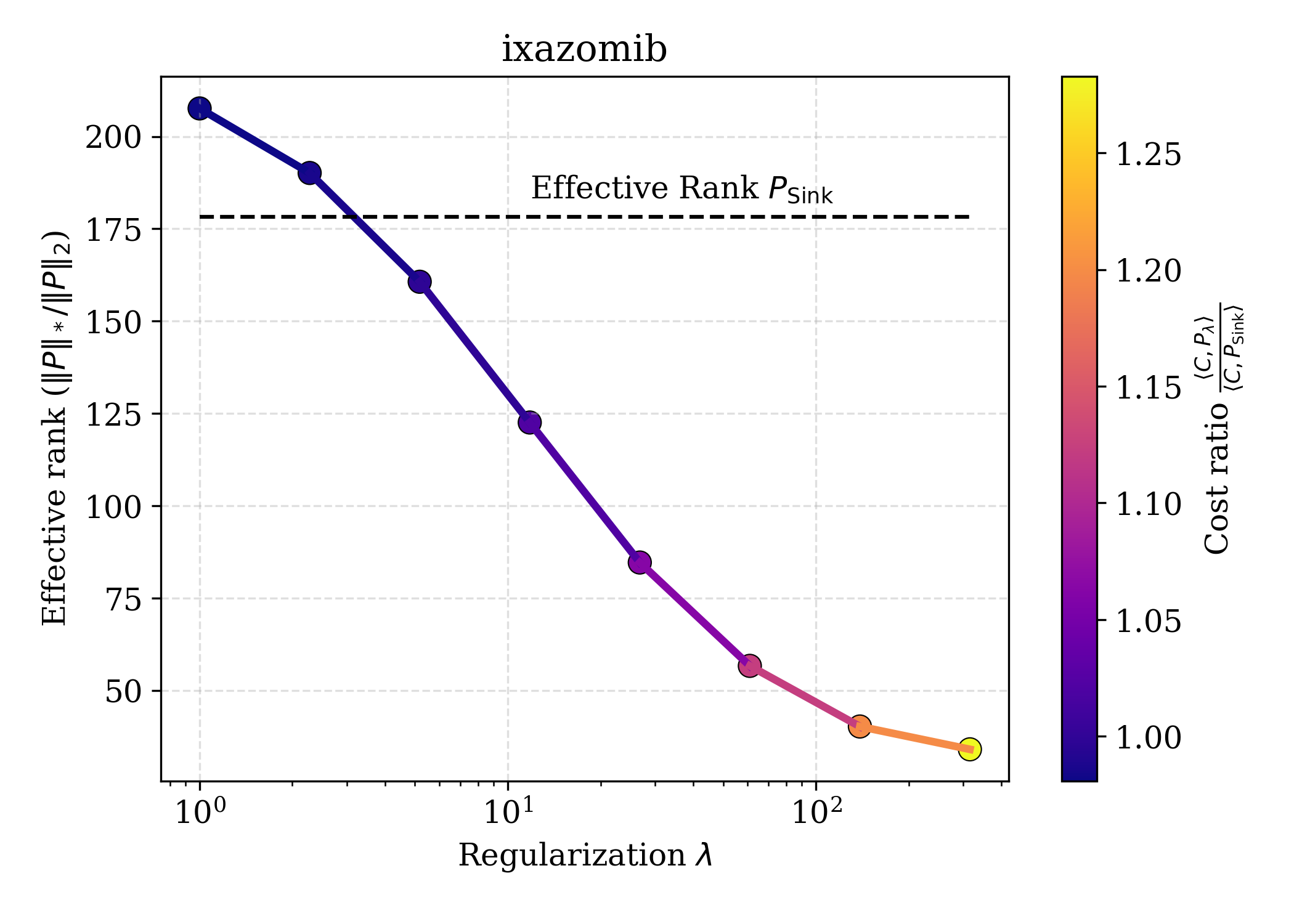}
    \includegraphics[width=0.35\linewidth]{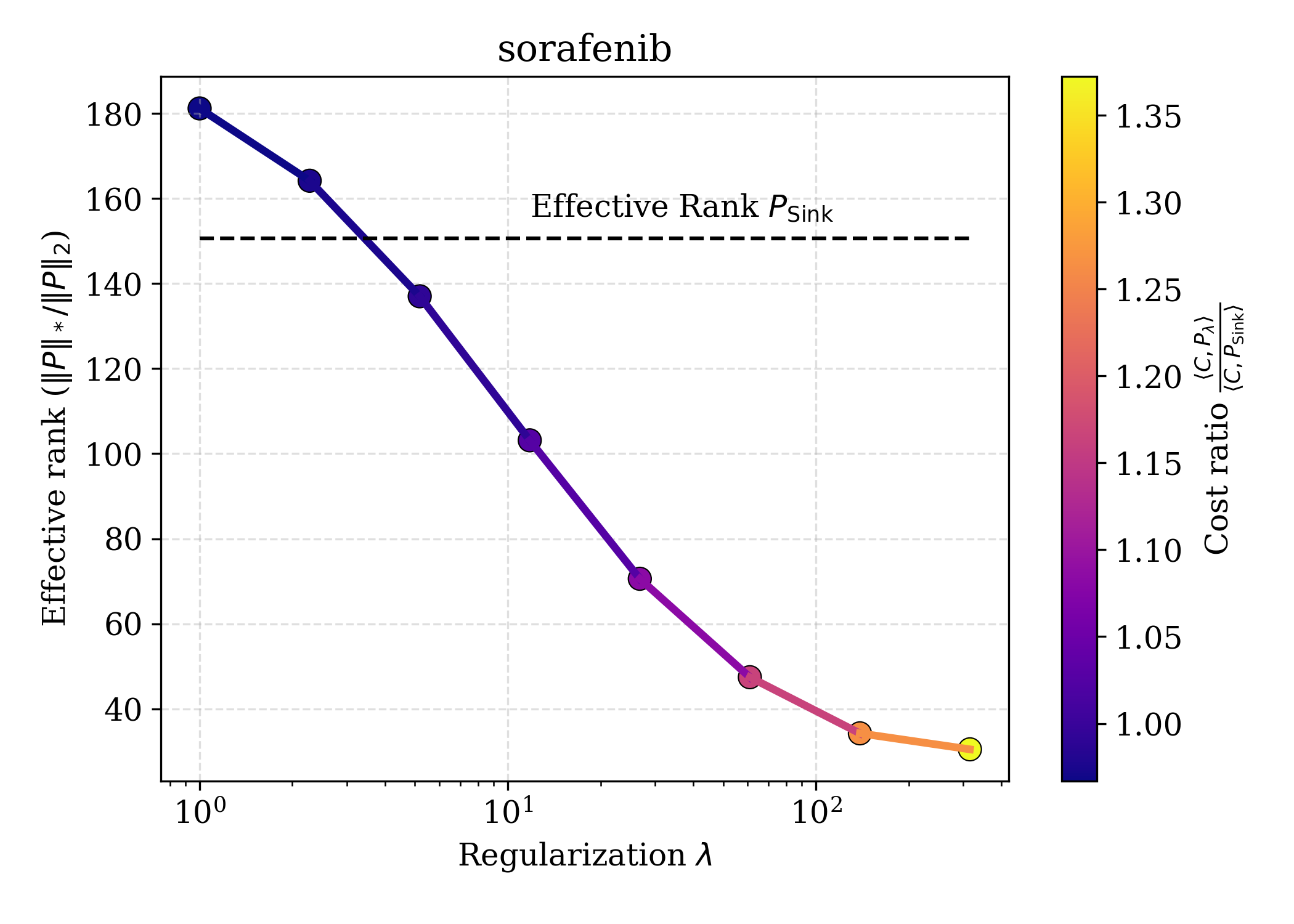}
    \includegraphics[width=0.35\linewidth]{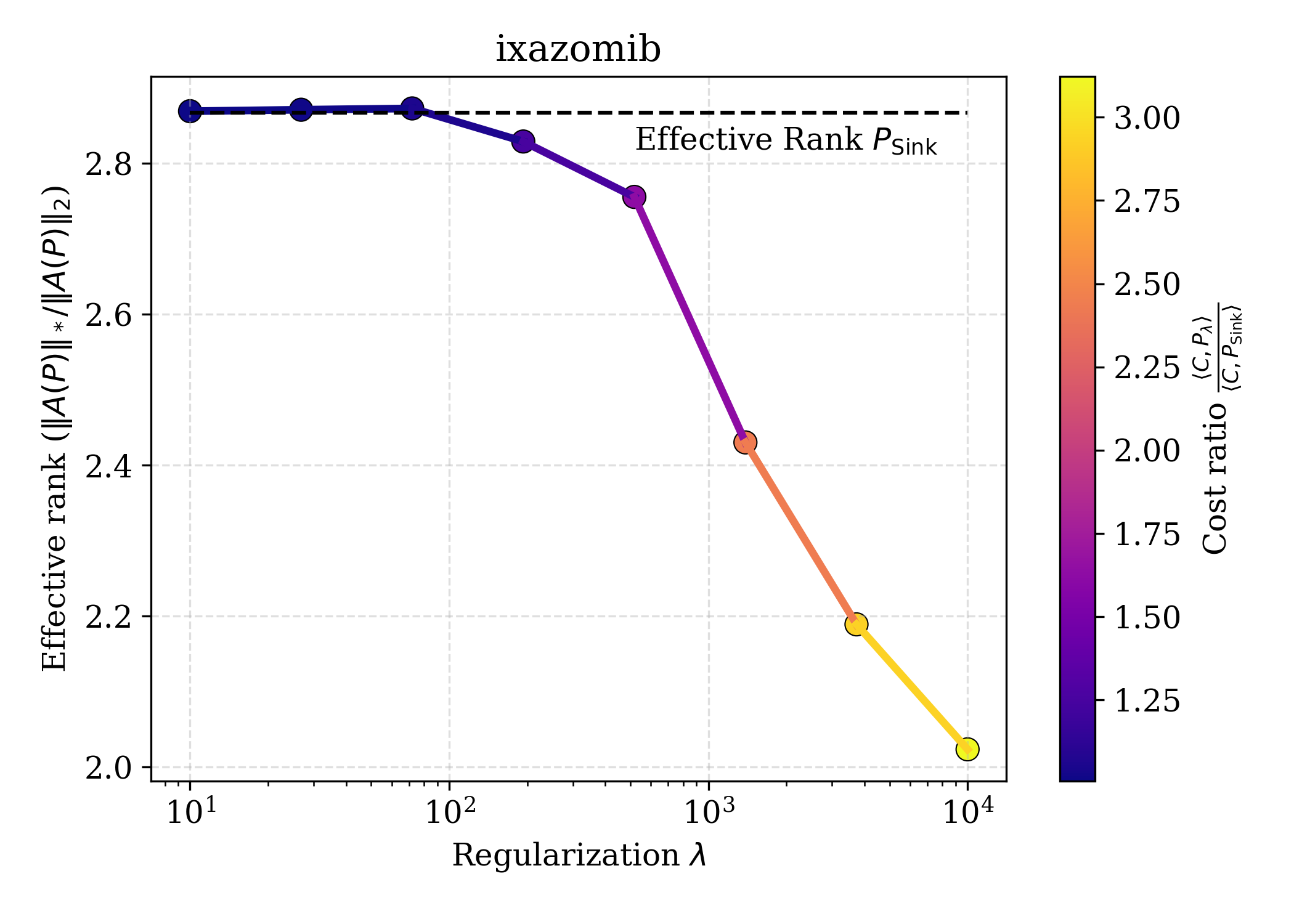}
    \includegraphics[width=0.35\linewidth]{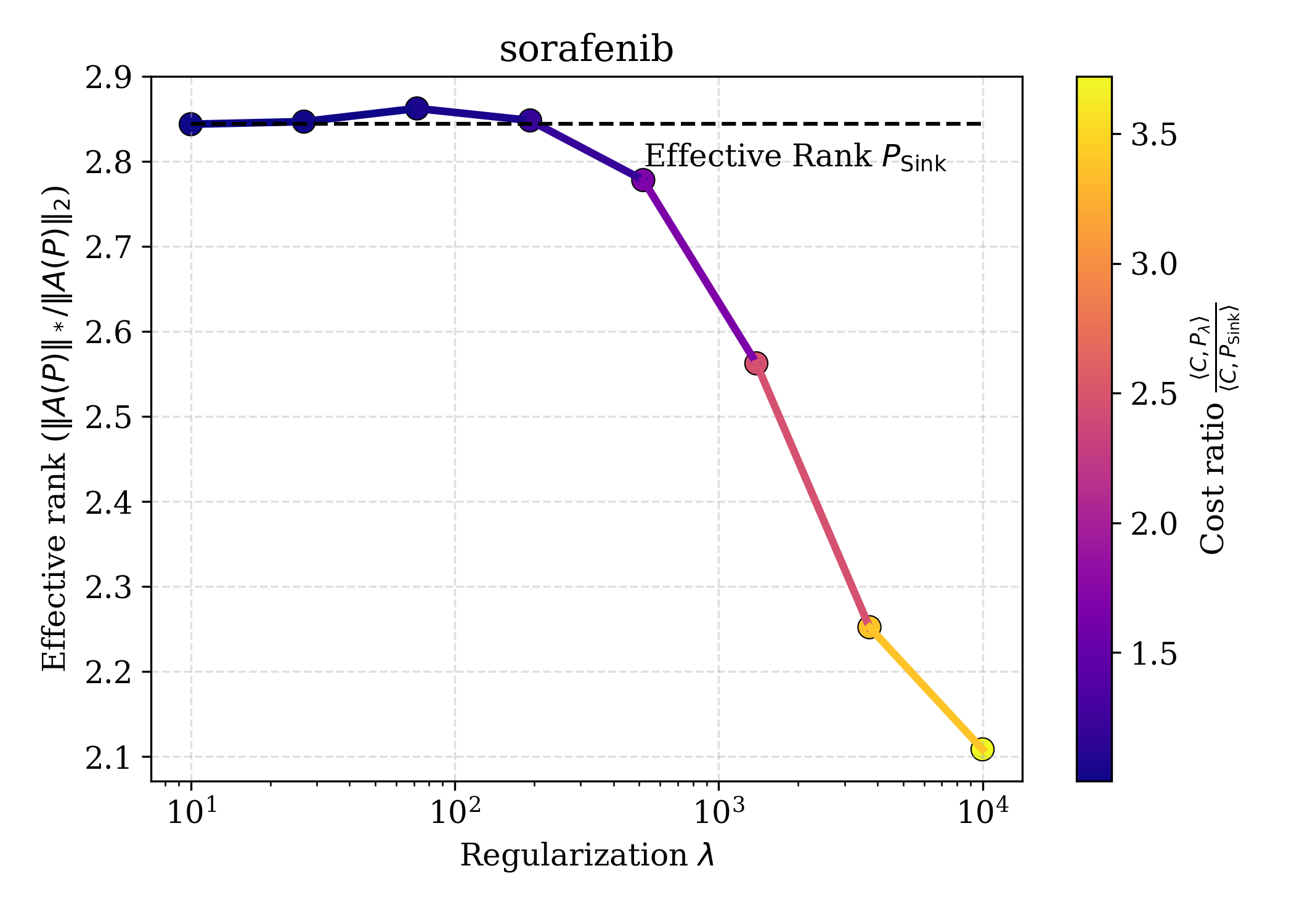}
    \caption{Plots of the performance of Schatten OT on the 4i perturbation data of  \cite{bunne2023learning}. For reference, we compare in all plots with what one gets using a Sinkhorn coupling with a regularization parameter of 1. Top: We examine the performance of Schatten-1 regularization in recovering a low-rank coupling for two different perturbations. As we can see, Schatten OT can reduce the effective rank while not increasing the transport cost too much. Bottom: We now show the performance of Schatten-1 regularization in recovering a low-rank barycentric projection map for two different perturbations. Again, Schatten OT can reduce the effective rank of this map, though the transport cost now increases more.}
    \label{fig:4i_lowrank}
\end{figure}

\section{Conclusion}

We introduced Schatten-$p$ regularized OT (Schatten OT), a unified convex framework for incorporating low-dimensional structure into OT problems. A key advantage of our formulation lies in its convexity and generality. Convexity allows us, for the first time, to provide provable recovery results in illustrative yet straightforward examples. Generality allows us to penalize any affine function of the coupling, thereby simultaneously encompassing many existing OT regularizations and enabling new ones. 

Theoretically, we established the first recovery guarantees for low-rank couplings and low-rank barycentric displacements, bridging ideas from compressed sensing and OT theory. 
Algorithmically, we developed an efficient mirror-descent method to solve these regularized problems in practice. Empirically, this approach performs well and demonstrates practical utility on 4i cell-perturbation data. Our results show that Schatten OT recovers low-rank structure with only modest increases in transport cost, yielding simpler and more interpretable transport maps.

We believe this work paves the way for more interpretable and scalable OT methods. In particular, the Schatten OT framework may provide a foundation for connecting OT to broader advances in sparse modeling, compressed sensing, and interpretable machine learning.


\bibliographystyle{plainnat}
\bibliography{refs}

\begin{thebibliography}{62}
\providecommand{\natexlab}[1]{#1}
\providecommand{\url}[1]{\texttt{#1}}
\expandafter\ifx\csname urlstyle\endcsname\relax
  \providecommand{\doi}[1]{doi: #1}\else
  \providecommand{\doi}{doi: \begingroup \urlstyle{rm}\Url}\fi

\bibitem[Altschuler et~al.(2017)Altschuler, Niles-Weed, and
  Rigollet]{altschuler2017near}
Jason Altschuler, Jonathan Niles-Weed, and Philippe Rigollet.
\newblock Near-linear time approximation algorithms for optimal transport via
  sinkhorn iteration.
\newblock \emph{Advances in neural information processing systems}, 30, 2017.

\bibitem[Alvarez-Melis et~al.(2019)Alvarez-Melis, Jegelka, and
  Jaakkola]{alvarez2019towards}
David Alvarez-Melis, Stefanie Jegelka, and Tommi~S Jaakkola.
\newblock Towards optimal transport with global invariances.
\newblock In \emph{The 22nd International Conference on Artificial Intelligence
  and Statistics}, pages 1870--1879. PMLR, 2019.

\bibitem[Arjovsky et~al.(2017)Arjovsky, Chintala, and
  Bottou]{arjovsky2017wasserstein}
Martin Arjovsky, Soumith Chintala, and L{\'e}on Bottou.
\newblock Wasserstein generative adversarial networks.
\newblock In \emph{International conference on machine learning}, pages
  214--223. PMLR, 2017.

\bibitem[Beck and Teboulle(2003)]{beck2003mirror}
Amir Beck and Marc Teboulle.
\newblock Mirror descent and nonlinear projected subgradient methods for convex
  optimization.
\newblock \emph{Operations Research Letters}, 31\penalty0 (3):\penalty0
  167--175, 2003.

\bibitem[Blondel et~al.(2018)Blondel, Seguy, and Rolet]{blondel2018smooth}
Mathieu Blondel, Vivien Seguy, and Antoine Rolet.
\newblock Smooth and sparse optimal transport.
\newblock In \emph{International conference on artificial intelligence and
  statistics}, pages 880--889. PMLR, 2018.

\bibitem[Bonneel and Digne(2023)]{bonneel2023survey}
Nicolas Bonneel and Julie Digne.
\newblock A survey of optimal transport for computer graphics and computer
  vision.
\newblock In \emph{Computer Graphics Forum}, volume~42, pages 439--460. Wiley
  Online Library, 2023.

\bibitem[Bubeck et~al.(2015)]{bubeck2015convex}
S{\'e}bastien Bubeck et~al.
\newblock Convex optimization: Algorithms and complexity.
\newblock \emph{Foundations and Trends{\textregistered} in Machine Learning},
  8\penalty0 (3-4):\penalty0 231--357, 2015.

\bibitem[Bunne et~al.(2023)Bunne, Stark, Gut, Del~Castillo, Levesque, Lehmann,
  Pelkmans, Krause, and R{\"a}tsch]{bunne2023learning}
Charlotte Bunne, Stefan~G Stark, Gabriele Gut, Jacobo~Sarabia Del~Castillo,
  Mitch Levesque, Kjong-Van Lehmann, Lucas Pelkmans, Andreas Krause, and Gunnar
  R{\"a}tsch.
\newblock Learning single-cell perturbation responses using neural optimal
  transport.
\newblock \emph{Nature methods}, 20\penalty0 (11):\penalty0 1759--1768, 2023.

\bibitem[Cai et~al.(2010)Cai, Cand{\`e}s, and Shen]{cai2010singular}
Jian-Feng Cai, Emmanuel~J Cand{\`e}s, and Zuowei Shen.
\newblock A singular value thresholding algorithm for matrix completion.
\newblock \emph{SIAM Journal on optimization}, 20\penalty0 (4):\penalty0
  1956--1982, 2010.

\bibitem[Cand{\`e}s and Tao(2010)]{candes2010power}
Emmanuel~J Cand{\`e}s and Terence Tao.
\newblock The power of convex relaxation: Near-optimal matrix completion.
\newblock \emph{IEEE transactions on information theory}, 56\penalty0
  (5):\penalty0 2053--2080, 2010.

\bibitem[Candes et~al.(2006)Candes, Romberg, and Tao]{candes2006stable}
Emmanuel~J Candes, Justin~K Romberg, and Terence Tao.
\newblock Stable signal recovery from incomplete and inaccurate measurements.
\newblock \emph{Communications on Pure and Applied Mathematics: A Journal
  Issued by the Courant Institute of Mathematical Sciences}, 59\penalty0
  (8):\penalty0 1207--1223, 2006.

\bibitem[Cand{\`e}s et~al.(2011)Cand{\`e}s, Li, Ma, and
  Wright]{candes2011robust}
Emmanuel~J Cand{\`e}s, Xiaodong Li, Yi~Ma, and John Wright.
\newblock Robust principal component analysis?
\newblock \emph{Journal of the ACM (JACM)}, 58\penalty0 (3):\penalty0 1--37,
  2011.

\bibitem[Chambolle and Pock(2011)]{chambolle2011first}
Antonin Chambolle and Thomas Pock.
\newblock A first-order primal-dual algorithm for convex problems with
  applications to imaging.
\newblock \emph{Journal of mathematical imaging and vision}, 40\penalty0
  (1):\penalty0 120--145, 2011.

\bibitem[Chen et~al.(2025)Chen, Xie, and Zhang]{chen2025displacement}
Peter Chen, Yue Xie, and Qingpeng Zhang.
\newblock Displacement-sparse neural optimal transport.
\newblock \emph{arXiv preprint arXiv:2502.01889}, 2025.

\bibitem[Chewi et~al.(2025)Chewi, Niles-Weed, and
  Rigollet]{chewi2025statistical}
Sinho Chewi, Jonathan Niles-Weed, and Philippe Rigollet.
\newblock \emph{Statistical optimal transport}.
\newblock \'{E}cole d'\'{E}t\'{e} de Probabilit\'{e}s de Saint-Flour. Springer,
  2025.

\bibitem[Cuturi(2013)]{cuturi2013sinkhorn}
Marco Cuturi.
\newblock Sinkhorn distances: Lightspeed computation of optimal transport.
\newblock \emph{Advances in neural information processing systems}, 26, 2013.

\bibitem[Cuturi et~al.(2023)Cuturi, Klein, and Ablin]{cuturi2023monge}
Marco Cuturi, Michal Klein, and Pierre Ablin.
\newblock Monge, bregman and occam: interpretable optimal transport in
  high-dimensions with feature-sparse maps.
\newblock In \emph{Proceedings of the 40th International Conference on Machine
  Learning}, pages 6671--6682, 2023.

\bibitem[Davis et~al.(2018)Davis, Drusvyatskiy, MacPhee, and
  Paquette]{davis2018subgradient}
Damek Davis, Dmitriy Drusvyatskiy, Kellie~J MacPhee, and Courtney Paquette.
\newblock Subgradient methods for sharp weakly convex functions.
\newblock \emph{Journal of Optimization Theory and Applications}, 179\penalty0
  (3):\penalty0 962--982, 2018.

\bibitem[Diamond and Boyd(2016)]{diamond2016cvxpy}
Steven Diamond and Stephen Boyd.
\newblock {CVXPY}: {A} {P}ython-embedded modeling language for convex
  optimization.
\newblock \emph{Journal of Machine Learning Research}, 17\penalty0
  (83):\penalty0 1--5, 2016.

\bibitem[Donoho(2006)]{donoho2006compressed}
David~L Donoho.
\newblock Compressed sensing.
\newblock \emph{IEEE Transactions on information theory}, 52\penalty0
  (4):\penalty0 1289--1306, 2006.

\bibitem[Eldar and Kutyniok(2012)]{eldar2012compressed}
Yonina~C Eldar and Gitta Kutyniok.
\newblock \emph{Compressed sensing: theory and applications}.
\newblock Cambridge university press, 2012.

\bibitem[Fan et~al.(2019)Fan, Ding, Chen, and Udell]{fan2019factor}
Jicong Fan, Lijun Ding, Yudong Chen, and Madeleine Udell.
\newblock Factor group-sparse regularization for efficient low-rank matrix
  recovery.
\newblock \emph{Advances in neural information processing Systems}, 32, 2019.

\bibitem[Fazel et~al.(2008)Fazel, Candes, Recht, and
  Parrilo]{fazel2008compressed}
Maryam Fazel, Emmanuel Candes, Ben Recht, and Pablo Parrilo.
\newblock Compressed sensing and robust recovery of low rank matrices.
\newblock In \emph{2008 42nd Asilomar Conference on Signals, Systems and
  Computers}, pages 1043--1047. IEEE, 2008.

\bibitem[Forrow et~al.(2019)Forrow, H{\"u}tter, Nitzan, Rigollet, Schiebinger,
  and Weed]{forrow2019statistical}
Aden Forrow, Jan-Christian H{\"u}tter, Mor Nitzan, Philippe Rigollet, Geoffrey
  Schiebinger, and Jonathan Weed.
\newblock Statistical optimal transport via factored couplings.
\newblock In \emph{The 22nd International Conference on Artificial Intelligence
  and Statistics}, pages 2454--2465. PMLR, 2019.

\bibitem[Gavish and Donoho(2017)]{gavish2017optimal}
Matan Gavish and David~L Donoho.
\newblock Optimal shrinkage of singular values.
\newblock \emph{IEEE Transactions on Information Theory}, 63\penalty0
  (4):\penalty0 2137--2152, 2017.

\bibitem[Genevay et~al.(2019)Genevay, Chizat, Bach, Cuturi, and
  Peyr{\'e}]{genevay2019sample}
Aude Genevay, L{\'e}naic Chizat, Francis Bach, Marco Cuturi, and Gabriel
  Peyr{\'e}.
\newblock Sample complexity of sinkhorn divergences.
\newblock In \emph{The 22nd international conference on artificial intelligence
  and statistics}, pages 1574--1583. PMLR, 2019.

\bibitem[Gonz{\'a}lez-Sanz and Nutz(2024)]{gonzalez2024sparsity}
Alberto Gonz{\'a}lez-Sanz and Marcel Nutz.
\newblock Sparsity of quadratically regularized optimal transport: Scalar case.
\newblock \emph{arXiv preprint arXiv:2410.03353}, 2024.

\bibitem[Halmos et~al.(2024)Halmos, Liu, Gold, and Raphael]{halmos2024low}
Peter Halmos, Xinhao Liu, Julian Gold, and Benjamin Raphael.
\newblock Low-rank optimal transport through factor relaxation with latent
  coupling.
\newblock \emph{Advances in Neural Information Processing Systems},
  37:\penalty0 114374--114433, 2024.

\bibitem[Halmos et~al.(2025)Halmos, Gold, Liu, and
  Raphael]{halmos2025hierarchical}
Peter Halmos, Julian Gold, Xinhao Liu, and Benjamin~J Raphael.
\newblock Hierarchical refinement: Optimal transport to infinity and beyond.
\newblock \emph{arXiv preprint arXiv:2503.03025}, 2025.

\bibitem[Huben et~al.(2024)Huben, Cunningham, Smith, Ewart, and
  Sharkey]{huben2024sparse}
Robert Huben, Hoagy Cunningham, Logan~Riggs Smith, Aidan Ewart, and Lee
  Sharkey.
\newblock Sparse autoencoders find highly interpretable features in language
  models.
\newblock In \emph{The Twelfth International Conference on Learning
  Representations}, 2024.
\newblock URL \url{https://openreview.net/forum?id=F76bwRSLeK}.

\bibitem[Jin et~al.(2021)Jin, Liu, and Xia]{jin2021two}
Kun Jin, Chaoyue Liu, and Cathy Xia.
\newblock Two-sided wasserstein procrustes analysis.
\newblock In \emph{IJCAI}, pages 3515--3521, 2021.

\bibitem[Kemertas et~al.(2025)Kemertas, Jepson, and
  Farahmand]{kemertas2025efficient}
Mete Kemertas, Allan~Douglas Jepson, and Amir-massoud Farahmand.
\newblock Efficient and accurate optimal transport with mirror descent and
  conjugate gradients.
\newblock \emph{Transactions on Machine Learning Research}, 2025.

\bibitem[Khamis et~al.(2024)Khamis, Tsuchida, Tarek, Rolland, and
  Petersson]{khamis2024scalable}
Abdelwahed Khamis, Russell Tsuchida, Mohamed Tarek, Vivien Rolland, and Lars
  Petersson.
\newblock Scalable optimal transport methods in machine learning: A
  contemporary survey.
\newblock \emph{IEEE transactions on pattern analysis and machine
  intelligence}, 2024.

\bibitem[Klein et~al.(2025)Klein, Palla, Lange, Klein, Piran, Gander,
  Meng-Papaxanthos, Sterr, Saber, Jing, et~al.]{klein2025mapping}
Dominik Klein, Giovanni Palla, Marius Lange, Michal Klein, Zoe Piran, Manuel
  Gander, Laetitia Meng-Papaxanthos, Michael Sterr, Lama Saber, Changying Jing,
  et~al.
\newblock Mapping cells through time and space with moscot.
\newblock \emph{Nature}, 638\penalty0 (8052):\penalty0 1065--1075, 2025.

\bibitem[Klein et~al.(2024)Klein, Pooladian, Ablin, Ndiaye, Niles-Weed, and
  Cuturi]{klein2024learning}
Michal Klein, Aram-Alexandre Pooladian, Pierre Ablin, Eug{\`e}ne Ndiaye,
  Jonathan Niles-Weed, and Marco Cuturi.
\newblock Learning elastic costs to shape monge displacements.
\newblock \emph{Advances in Neural Information Processing Systems},
  37:\penalty0 108542--108565, 2024.

\bibitem[Liao and Gu(2025)]{liao2025arcagiwithoutpretraining}
Isaac Liao and Albert Gu.
\newblock Arc-agi without pretraining, 2025.
\newblock URL
  \url{https://iliao2345.github.io/blog_posts/arc_agi_without_pretraining/arc_agi_without_pretraining.html}.

\bibitem[Lin et~al.(2021)Lin, Azabou, and Dyer]{lin2021making}
Chi-Heng Lin, Mehdi Azabou, and Eva~L Dyer.
\newblock Making transport more robust and interpretable by moving data through
  a small number of anchor points.
\newblock \emph{Proceedings of machine learning research}, 139:\penalty0 6631,
  2021.

\bibitem[Lorenz et~al.(2021)Lorenz, Manns, and Meyer]{lorenz2021quadratically}
Dirk~A Lorenz, Paul Manns, and Christian Meyer.
\newblock Quadratically regularized optimal transport.
\newblock \emph{Applied Mathematics \& Optimization}, 83\penalty0 (3):\penalty0
  1919--1949, 2021.

\bibitem[Lu et~al.(2015)Lu, Tang, Yan, and Lin]{lu2015nonconvex}
Canyi Lu, Jinhui Tang, Shuicheng Yan, and Zhouchen Lin.
\newblock Nonconvex nonsmooth low rank minimization via iteratively reweighted
  nuclear norm.
\newblock \emph{IEEE Transactions on Image Processing}, 25\penalty0
  (2):\penalty0 829--839, 2015.

\bibitem[Maunu et~al.(2019)Maunu, Zhang, and Lerman]{maunu2019well}
Tyler Maunu, Teng Zhang, and Gilad Lerman.
\newblock A well-tempered landscape for non-convex robust subspace recovery.
\newblock \emph{Journal of Machine Learning Research}, 20\penalty0
  (37):\penalty0 1--59, 2019.

\bibitem[Nemirovsky and Yudin(1983)]{nemirovsky1983problem}
Arkadij~Semenovi{\v{c}} Nemirovsky and David~Borisovich Yudin.
\newblock Problem complexity and method efficiency in optimization.
\newblock 1983.

\bibitem[Nesterov and Nemirovski(2013)]{nesterov2013first}
Yurii Nesterov and Arkadi Nemirovski.
\newblock On first-order algorithms for l1/nuclear norm minimization.
\newblock \emph{Acta Numerica}, 22:\penalty0 509--575, 2013.

\bibitem[Nie et~al.(2012)Nie, Huang, and Ding]{nie2012low}
Feiping Nie, Heng Huang, and Chris Ding.
\newblock Low-rank matrix recovery via efficient schatten p-norm minimization.
\newblock In \emph{Proceedings of the AAAI Conference on Artificial
  Intelligence}, volume~26, pages 655--661, 2012.

\bibitem[Niles-Weed and Rigollet(2022)]{niles2022estimation}
Jonathan Niles-Weed and Philippe Rigollet.
\newblock Estimation of wasserstein distances in the spiked transport model.
\newblock \emph{Bernoulli}, 28\penalty0 (4):\penalty0 2663--2688, 2022.

\bibitem[Paty and Cuturi(2019)]{paty2019subspace}
Fran{\c{c}}ois-Pierre Paty and Marco Cuturi.
\newblock Subspace robust wasserstein distances.
\newblock In \emph{International conference on machine learning}, pages
  5072--5081. PMLR, 2019.

\bibitem[Peyr{\'e} and Cuturi(2019)]{peyre2019computational}
Gabriel Peyr{\'e} and Marco Cuturi.
\newblock Computational optimal transport: With applications to data science.
\newblock \emph{Foundations and Trends{\textregistered} in Machine Learning},
  11\penalty0 (5-6):\penalty0 355--607, 2019.

\bibitem[Recht et~al.(2010)Recht, Fazel, and Parrilo]{recht2010guaranteed}
Benjamin Recht, Maryam Fazel, and Pablo~A Parrilo.
\newblock Guaranteed minimum-rank solutions of linear matrix equations via
  nuclear norm minimization.
\newblock \emph{SIAM review}, 52\penalty0 (3):\penalty0 471--501, 2010.

\bibitem[Scarvelis and Solomon(2024)]{scarvelis2024nuclear}
Christopher Scarvelis and Justin~M Solomon.
\newblock Nuclear norm regularization for deep learning.
\newblock \emph{Advances in Neural Information Processing Systems},
  37:\penalty0 116223--116253, 2024.

\bibitem[Scetbon et~al.(2021)Scetbon, Cuturi, and Peyr{\'e}]{scetbon2021low}
Meyer Scetbon, Marco Cuturi, and Gabriel Peyr{\'e}.
\newblock Low-rank sinkhorn factorization.
\newblock In \emph{International Conference on Machine Learning}, pages
  9344--9354. PMLR, 2021.

\bibitem[Schiebinger et~al.(2019)Schiebinger, Shu, Tabaka, Cleary, Subramanian,
  Solomon, Gould, Liu, Lin, Berube, et~al.]{schiebinger2019optimal}
Geoffrey Schiebinger, Jian Shu, Marcin Tabaka, Brian Cleary, Vidya Subramanian,
  Aryeh Solomon, Joshua Gould, Siyan Liu, Stacie Lin, Peter Berube, et~al.
\newblock Optimal-transport analysis of single-cell gene expression identifies
  developmental trajectories in reprogramming.
\newblock \emph{Cell}, 176\penalty0 (4):\penalty0 928--943, 2019.

\bibitem[Schr{\"o}dinger(1932)]{schrodinger1932theorie}
Erwin Schr{\"o}dinger.
\newblock Sur la th{\'e}orie relativiste de l'{\'e}lectron et
  l'interpr{\'e}tation de la m{\'e}canique quantique.
\newblock In \emph{Annales de l'institut Henri Poincar{\'e}}, volume~2, pages
  269--310, 1932.

\bibitem[Sebbouh et~al.(2024)Sebbouh, Cuturi, and
  Peyr{\'e}]{sebbouh2024structured}
Othmane Sebbouh, Marco Cuturi, and Gabriel Peyr{\'e}.
\newblock Structured transforms across spaces with cost-regularized optimal
  transport.
\newblock In \emph{International Conference on Artificial Intelligence and
  Statistics}, pages 586--594. PMLR, 2024.

\bibitem[Sinkhorn(1967)]{sinkhorn1967diagonal}
Richard Sinkhorn.
\newblock Diagonal equivalence to matrices with prescribed row and column sums.
\newblock \emph{The American Mathematical Monthly}, 74\penalty0 (4):\penalty0
  402--405, 1967.

\bibitem[Srebro et~al.(2004)Srebro, Rennie, and Jaakkola]{srebro2004maximum}
Nathan Srebro, Jason Rennie, and Tommi Jaakkola.
\newblock Maximum-margin matrix factorization.
\newblock \emph{Advances in neural information processing systems}, 17, 2004.

\bibitem[Villani et~al.(2008)]{villani2008optimal}
C{\'e}dric Villani et~al.
\newblock \emph{Optimal transport: old and new}, volume 338.
\newblock Springer, 2008.

\bibitem[Wright and Ma(2022)]{wright2022high}
John Wright and Yi~Ma.
\newblock \emph{High-dimensional data analysis with low-dimensional models:
  Principles, computation, and applications}.
\newblock Cambridge University Press, 2022.

\bibitem[Xie et~al.(2016)Xie, Gu, Liu, Zuo, Zhang, and Zhang]{xie2016weighted}
Yuan Xie, Shuhang Gu, Yan Liu, Wangmeng Zuo, Wensheng Zhang, and Lei Zhang.
\newblock Weighted schatten $ p $-norm minimization for image denoising and
  background subtraction.
\newblock \emph{IEEE transactions on image processing}, 25\penalty0
  (10):\penalty0 4842--4857, 2016.

\bibitem[Yu et~al.(2020)Yu, Chan, You, Song, and Ma]{yu2020learning}
Yaodong Yu, Kwan Ho~Ryan Chan, Chong You, Chaobing Song, and Yi~Ma.
\newblock Learning diverse and discriminative representations via the principle
  of maximal coding rate reduction.
\newblock \emph{Advances in neural information processing systems},
  33:\penalty0 9422--9434, 2020.

\bibitem[Yu et~al.(2023)Yu, Buchanan, Pai, Chu, Wu, Tong, Haeffele, and
  Ma]{yu2023white}
Yaodong Yu, Sam Buchanan, Druv Pai, Tianzhe Chu, Ziyang Wu, Shengbang Tong,
  Benjamin Haeffele, and Yi~Ma.
\newblock White-box transformers via sparse rate reduction.
\newblock \emph{Advances in Neural Information Processing Systems},
  36:\penalty0 9422--9457, 2023.

\bibitem[Yuan and Yang(2009)]{yuan2009sparse}
Xiaoming Yuan and Junfeng Yang.
\newblock Sparse and low-rank matrix decomposition via alternating direction
  methods.
\newblock \emph{preprint}, 12\penalty0 (2), 2009.

\bibitem[Yurtsever et~al.(2021)Yurtsever, Tropp, Fercoq, Udell, and
  Cevher]{yurtsever2021scalable}
Alp Yurtsever, Joel~A Tropp, Olivier Fercoq, Madeleine Udell, and Volkan
  Cevher.
\newblock Scalable semidefinite programming.
\newblock \emph{SIAM Journal on Mathematics of Data Science}, 3\penalty0
  (1):\penalty0 171--200, 2021.

\bibitem[Zhang et~al.(2018)Zhang, Wei, and Yang]{zhang2018learning}
Yu~Zhang, Ying Wei, and Qiang Yang.
\newblock Learning to multitask.
\newblock \emph{Advances in Neural Information Processing Systems}, 31, 2018.

\end{thebibliography}

\appendix

\section{Extension to the Continuous Setting}
\label{app:cts_extension}

Up until now, we have focused our attention on formulations in the discrete case. However, there is a direct extension of Schatten OT to the continuous setting by taking Schatten-$p$ norms of appropriate linear operators over general Hilbert spaces. In this section, we let $\mu, \nu \in \cP_{2}(\R^d)$ be general measures. The set of couplings between these measures is $\Pi(\mu, \nu)$. 

To define our extension to the continuous case, we let $\cA:\cP_2(\R^d \times \R^d)\to \cB(\cH)$ be a map from the space of couplings to the set of bounded linear operators on some Hilbert space $\cH$. Let $\|\cdot\|_{S_p}$ now denote the Schatten-$p$ norm over $\cB(\cH)$, which is defined as $\|T\|_{S_p}^p = \Tr[(T^* T)^{p/2}]$. Then, we define the continuous Schatten OT problem 
\begin{equation}\label{eq:schatten_ot_cts}  
    \SOT_p(\mu, \nu; (\lambda, p, q, \cA)) := \min_{\pi \in \Pi(\mu, \nu)} \E_{(X,Y) \sim \pi} \|X-Y\|^2 + \lambda \|\cA(\pi)\|_{S_p}^q.
\end{equation}
We note that, as in the discrete case, this notion depends on choices of $\lambda$, $p$, $q$, and $\cA$. 
As before, choosing $p,q \geq 1$ and $\cA$ an affine map makes the problem \eqref{eq:schatten_ot_cts} convex.

Below, we give some examples of affine maps that extend our discrete examples. Throughout, we let $\rho = \mu \otimes \nu$ be the reference product measure.

\paragraph{Covariance regularization}
The most direct connection between the continuous and discrete cases is to penalize moments of the distribution. In the continuous case, this corresponds to regularizing the covariance of $\pi$. In this case, all of the regularizations discussed in Section \ref{subsec:schatten_ot} are the same except we regularize the linear operator over $\R^d$,
$\bSigma_{\pi} = \E_{\pi} \begin{pmatrix}
        X \\Y
    \end{pmatrix}\begin{pmatrix}
        X \\Y
    \end{pmatrix}^\top.$

\paragraph{Continuous sparse and low-rank regularization:} We now discuss the analogs of quadratic and low-rank regularization, the coupling matrix $\bP$. Here, we can take $\cA(\pi) = S_{\pi}:L^2(\nu) \to L^2(\mu)$ as the linear operator 
\[
    (S_\pi f)(\bx) = \int f(y) \frac{d \pi}{d \rho} d\nu(y) = \E_{\pi} [f(Y)|X=\bx].
\]
Note that $T_\pi$ is affine in $\pi$. Then, regularizing the Schatten-$p$ norm of $S_\pi$ corresponds to using the Schatten-$p$ norm of $\bP$ as discussed earlier. The continuous quadratic case, where the continuous case has already been studied \cite{lorenz2021quadratically}, but not as a Schatten-$2$ norm of the operator $S_\pi$. 

\paragraph{Elastic costs:} We can recover the elastic costs by taking the Schatten norm of a specifically constructed operator. Choose a measurable partition of $\R^d \times \R^d$ given by $(E_k)_{k \in \N}$ with $\rho(E_k) > 0$. Define an orthonormal family in $L^2(\rho)$ by 
\[
    \phi_k(\bx, \by) = \frac{\mathbbm{1}((\bx,\by) \in E_k)}{\sqrt{\rho(E_k)}}
\]
Set the diagonal weights to be $s_k(\pi) = \int_{E_k} \|\by - \bx\|_1 d\pi$, which are linear in $\pi$. Then, letting $(e_k)_{k \in \N}$ be a basis for $\ell_2$, we can  define the linear operator $\cA(\pi):\ell^2 \to L^2(\rho)$ by
\[
    A(\pi)e_k = s_k(\phi) \phi_k.
\]
Notice that this is again linear in $\pi$, and furthermore $A(\pi)e_k$ are orthogonal with $\|A(\pi)e_k\| = s_n(\pi)$. Therefore, the singular values of $\cA(\pi)$ are $s_n(\pi)$. Thus
\[
    \|\cA(\pi)\|_{S_1} = \sum_k s_k(\pi) = \int \|\by-\bx\|_1 d\pi.
\]
A similar construction yields the subspace elastic costs discussed in Section \ref{subsec:schatten_ot}. The general principle here is that elastic OT problems can be embedded as Schatten OT regularized problems over appropriate operators.

\paragraph{Barycentric projection maps and displacements:} We can also consider Schatten-$p$ regularization of the barycentric projection maps and displacements. In the continuous case, the barycentric projection map is $T_{\pi}(\cdot) = \E_{\pi(Y|\cdot)} Y$. Let the displacement map be $D_\pi(\cdot) = \E_{\pi(Y|\cdot)} Y - \cdot$. Then, we can formulate a Schatten-$p$ penalization of this  barycentric displacement map by viewing $T_{\pi}$ or $D_{\pi}$ as operators $T_{\pi},D_{\pi}:\R^d \to L^2(\mu)$ given by $T_{\pi} v = \langle v, T_\pi(\cdot) \rangle$ and $D_{\pi} v = \langle v, D_\pi(\cdot) \rangle$.

\section{Supplementary Proofs}

\subsection{Proof of Theorem \ref{thm:R-vs-R}}

\begin{proof}
Define, for each $t$, the cluster indicator mass vectors
\[
\balpha^{(t)}\in\R^n, (\balpha^{(t)})_i=
\begin{cases}
\frac{1}{Rg}, & i\in S_t,\\
0, & \text{otherwise},
\end{cases}
\]
and similarly
\[
\bbeta^{(t)}\in\R^m, (\bbeta^{(t)})_j=
\begin{cases}
\frac{1}{Rg}, & j\in T_t,\\
0, & \text{otherwise}.
\end{cases}
\]
The rank $R$ coupling we wish to recover is
\begin{equation}\label{eq:Pstar}
\bP^\star\ :=\ \sum_{t=1}^R \balpha^{(t)} {\bbeta^{(t)}}^\top.
\end{equation}
Notice that $\bP^\star\in \cU(\ba, \bb)$ is block-diagonal with blocks $(S_t\times T_t)$ that are uniform (each entry equals $1/(Rg)^2$), and we can explicitly compute $\|\bP^\star\|_{S_1}= \frac{1}{R}.$ It will be convenient to define the normalized indicator vectors $\bu^{(t)}=\balpha^{(t)}/\|\balpha^{(t)}\|_2$, $\bv^{(t)}=\bbeta^{(t)}/\|\bbeta^{(t)}\|_2$, which we stack into matrices $\bU^\star=[\bu^{(1)},\dots,\bu^{(R)}]$, $\bV^\star=[\bv^{(1)},\dots,\bv^{(R)}]$. In this way, the canonical subgradient of $\|\cdot\|_{S_1}$ at $\bP^\star$ is $\bG^\star:=\bU^\star {\bV^\star}^\top$.

\textbf{Step 1: Lower bound on $\Delta_{\min}$.}
For any $\bx\in B(\bc_t,\rho)$, $\by_{\mathrm{in}}\in B(\bd_t,\rho)$, $\by_{\mathrm{out}}\in B(\bd_s,\rho)$ with $s\neq t$,
\[
\|\bx-\by_{\mathrm{out}}\| \geq \|\bc_t-\bd_s\| - \|\bx-\bc_t\| - \|\by_{\mathrm{out}}-\bd_s\| \geq \Gamma-2\rho,
\]
and
\[
\|\bx-\by_{\mathrm{in}}\| \leq \|\bc_t-\bd_t\| + \|\bx-\bc_t\| + \|\by_{\mathrm{in}}-\bd_t\| \leq \gamma+2\rho.
\]
Thus
\[
\|\bx-\by_{\mathrm{out}}\|^2 - \|\bx-\by_{\mathrm{in}}\|^2 \geq (\Gamma-2\rho)^2 - (\gamma+2\rho)^2,
\]
which is positive when \eqref{eq:sep-margin} holds.

\textbf{Step 2: Across-block exclusion via tilted cost.} By convexity of $\|\cdot\|_{S_1}$,
\[
\|\bP\|_{S_1} \geq \|\bP^\star\|_{S_1} + \langle \bG^\star, \bP-\bP^\star\rangle,
  \bG^\star\in\partial\|\bP^\star\|_{S_1},\ \bG^\star=\bU^\star {\bV^\star}^\top.
\]
Hence, for any feasible $P$,
\[
\langle \bC, \bP\rangle+\lambda\|\bP\|_{S_1} - (\langle \bC,\bP^\star\rangle+\lambda\|\bP^\star\|_{S_1})
 \geq \big\langle \bS(\lambda, \bG^\star), \bP-\bP^\star\big\rangle.
\]
For $i\in S_t$ and $j\in T_s$ with $s\neq t$, one has $\bG^\star_{ij}=0$, since the left and right singular vectors are block-supported and orthonormal across clusters. On the other hand, for $j'\in T_t$,
\[
\bG^\star_{i j'} = \langle \bu^{(t)},\be_i\rangle \langle \bv^{(t)},\be_{j'}\rangle
=\frac{a_i}{\|\balpha^{(t)}\|_2}\cdot \frac{b_{j'}}{\|\bbeta^{(t)}\|_2} = \frac{1}{g}.
\]
Therefore, for any $i\in S_t$, $s\neq t$, $j\in T_s$, and $j'\in T_t$,
\[
\bS_{ij}(\lambda, \bG^\star)-\bS_{i j'}(\lambda, \bG^\star)
=(\|\bx_i-\by_j\|_2^2 - \|\bx_i-\by_{j'}\|_2^2) - \lambda\cdot \frac{1}{g}
\ \ge\ \Delta_{\min} - \frac{\lambda}{g}.
\]
If $\lambda<g\Delta_{\min}$, these gaps are strictly positive, so no $S(\lambda, \bG^\star)$-optimal coupling can place mass across clusters. Any minimizer of the original problem must then be block-supported on $\bigcup_t (S_t\times T_t)$.

\smallskip
\textbf{Step 3: Within-block tie-breaking via the nuclear norm.} By the distance equality condition in Assumption \ref{assump:separation}, all within-cluster couplings have equal transport cost. Fix $t$. We can restrict any feasible coupling $\bP \in \cU(\ba, \bb)$ to the block $(S_t,T_t)$, which we denote as $\bP_{S_t,T_t}$. We note that this can be written as
\[
\bP_{S_t,T_t}=\bone_g \bone_g^\top / g^2 + \bM^{(t)}, \text{ where }
  \bM^{(t)}\mathbf{1}=\bzero,  (\bM^{(t)})^\top\mathbf{1}=\bzero.
\]
In other words, we can represent it as rank-1 product coupling plus a perturbation with $\bzero$ row/column sums. Choose an orthonormal basis of $\R^{g}$ on the target side with the first vector proportional to $\bone_g$. Then $\bM^{(t)}$ lives entirely in the orthogonal complement of $\bone_g$. The standard inequality $\|\cdot\|_{S_1}\geq \|\cdot\|_{S_2}$ yields
\[
\|\bP_{S_t,T_t}\|_{S_1} \geq \|\bP_{S_t,T_t}\|_{S_2}
=\sqrt{\|\bone_g \bone_g^\top / g^2 \|_{S_2}^2 + \|\bM^{(t)}\|_{S_2}^2}
> \|\bone_g \bone_g^\top / g^2 \|_{S_2}
=\|\bone_g/g\|_2 \|\bone_g/g\|_2,
\]
whenever $\bM^{(t)}\neq 0$. Summing over $t$ shows that among all block-supported couplings, the nuclear norm is uniquely minimized at $\bM^{(t)}\equiv 0$, i.e., at the uniform block $\bP^\star$.

Combining these three steps proves the proposition.
\end{proof}

\subsection{Proof of Theorem \ref{thm:rank1-recovery}}

\begin{proof}
We proceed in four steps. We will show that the coupling we recover, $\bP^{\star}\in\Pi(\ba, \bb)$, satisfies the within-cluster equal split condition given by
$
\bP^{\star}_{i,(t,+)}\ =\ \bP^{\star}_{i,(t,-)}\ =\ \tfrac12 a_i$ if $i\in S_t$ otherwise $\bP^{\star}_{i,(s,\sigma)} = 0$ if $s\ne t$ or $i\notin S_t$.

\medskip
\noindent\textbf{Step 1: Feasibility and rank-$1$ structure.}
By construction, $\bP^{\star}\in \cU(\ba, \bb)$ and, for each $i\in S_t$,
\[
T_{\bP^{\star}}(\bx_i)=\tfrac12(\by_{t,+}+\by_{t,-})=\bm_t.
\]
Hence $T_{\bP^{\star}}(\bx_i) - \bx_i=\bm_t-\bx_i = -\xi_i \bu$ when $\bx_i=\bm_t+\xi_i \bu$ for $|\xi_i|\le\rho$,  which is true by assumption. Writing $\bgamma\in\R^n$ such that $  \gamma_i= - \xi_i \sqrt{a_i}$, we have
\[
\cA (\bP^{\star}) = \bu \bgamma^\top.
\]
Therefore, $\rank \cA (\bP^{\star})=1$, and $\|\cA (\bP^{\star})\|_{S_1} = \|\gamma\|$.

\medskip
\noindent\textbf{Step 2: A tilted-cost lower bound and across-cluster margin.}
Let $\bG^\star$ be a canonical subgradient of the nuclear norm at $\bB^\star := \cA(\bP^\star)$:
\[
\bG^\star\ \in\ \partial \|\bB^\star\|_{S_1},  
\bG^\star  =  \bu \bw^\top,\text{ where }  \bw:=\frac{\bgamma}{\|\bgamma\|_2}.
\]
For any $\bP\in\cU(\ba, \bb)$, by convexity of the nuclear norm,
\begin{equation}\label{eq:subgrad-ineq}
\|\cA(\bP)\|_{S_1}  \geq  \|\bB^\star\|_{S_1}  +  \langle \bG^\star, \cA(\bP)-\bB^\star \rangle.
\end{equation}
Using $\cA(\bP)-\bB^\star = \bY(\bP-\bP^{\star})^\top \bA^{-1/2}$ and cyclicity of the trace,
\begin{equation}\label{eq:cyclic_tr}
\langle \bG^\star, \cA(\bP)-\bB^\star \rangle
 =  \langle \bA^{-1/2}{\bG^\star}^\top \bY, \bP-\bP^{\star}\rangle.
\end{equation}

Combining~\eqref{eq:subgrad-ineq} and \eqref{eq:cyclic_tr} with the objective $F_\lambda(\bP):=\langle \bC, \bP\rangle+\lambda\|\cA(\bP)\|_{S_1}$ and the definition of the tilted cost $\bS(\lambda, \bG)$ in \eqref{eq:S-def} yields the lower bound
\begin{equation}\label{eq:fund-lb}
F_\lambda(\bP)-F_\lambda \left(\bP^{\star}\right)
 \geq  \langle \bS(\lambda, \bG^\star), \bP-\bP^{\star}\rangle.
\end{equation}

We can compute the tilted costs $\bS(\lambda, \bG^\star)$ explicitly: for any $i$ and $(t, \sigma)$,
\[
(\bA^{-1/2}{\bG^\star}^\top \bY)_{i,(t,\sigma)}
 =  \frac{1}{\sqrt{a_i}} w_i \langle \bu, \by_{t,\sigma}\rangle
 = \frac{\gamma_i \mu_t}{\|\bgamma\|_2\sqrt{a_i}}
 =  - \frac{\xi_i \mu_t}{\|\gamma\|_2}.
\]
Therefore, assuming that $i \in S_t$ and for any $s \neq t$, $\sigma\in\{\pm\}$,
\begin{align}
\bS_{i,(s,\sigma)}(\lambda, \bG^\star)-\bS_{i,(t,\pm)}(\lambda, \bG^\star)
&=\underbrace{\|x_i-y_{s,\sigma}\|^2-\|x_i-y_{t,\pm}\|^2}_{=\Delta_{i,s}}
 + \lambda\left(-\frac{\xi_i}{\|\gamma\|_2}\right)(\mu_s-\mu_t)
\label{eq:S-diff}
\end{align}
By assumption,
\begin{equation}\label{eq:Delta-lb}
\Delta_{i,s}\ \ge\ |\mu_t-\mu_s|(|\mu_t-\mu_s|-2\rho)\ \ge\ \Lambda(\Lambda-2\rho)\ >\ 0.
\end{equation}
Also, since $|\xi_i|\leq\rho$, we can bound $\|\gamma\|_2^2  = \sum_{k=1}^n a_k \xi_k^2  \leq  \rho^2\sum_{k=1}^n a_k  = \rho^2$. This implies that $\frac{|\xi_i|}{\|\gamma\|_2}\ \le\ 1$.
Thus, we can extend the lower bound in~\eqref{eq:S-diff} 
\begin{equation}\label{eq:reduced-cost-gap}
\bS_{i,(s,\sigma)}(\lambda,\bG^\star)-\bS_{i,(t,\pm)}(\lambda, \bG^\star)
 \geq \Lambda(\Lambda-2\rho)-\lambda).
\end{equation}
Thus, at $\bG^\star$, for every $\lambda\in[0,\Lambda-2\rho)$, across-cluster tilted costs are strictly greater than within-cluster tilted costs. Therefore any tilted cost optimal coupling must match $\bx_i$ to $\{\by_{t, \pm}\}$ for $i \in S_t$.

\medskip
\noindent\textbf{Step 3: Within-cluster degeneracy and the nuclear-norm tie-break.}
Fix $t\in[R]$. For $i\in S_t$, any within-cluster move between the symmetric targets $(t,+)$ and $(t,-)$ has zero cost difference, 
\[
\|\bx_i-\by_{t,+}\|^2=\|\bx_i-\by_{t,-}\|^2.
\]
Moreover, the tilted cost is the same for $(t,+)$ and $(t,-)$, since $\langle \bu,\by_{t,+}\rangle=\langle \bu,\by_{t,-}\rangle=\mu_t$.
Therefore, for any feasible $\bP$ that sends mass within clusters (i.e., $\supp(\bP)\subseteq\{(i,(t,\pm)):\ i\in S_t\}$),
\begin{equation}\label{eq:within-cluster-Szero}
\langle \bS(\lambda, \bG^\star),\bP-\bP^{\star}\rangle = 0.
\end{equation}

For such $\bP$, we can write the within-cluster mass split by $p_i\in[0,1]$ so that
\[
\bP_{i,(t,+)}= p_i a_i,\ \bP_{i,(t,-)}=(1-p_i)a_i
\]
A direct computation gives
\[
T_{\bP}(\bx_i)= p_i \by_{t,+} + (1-p_i)\by_{t,-}
= \bm_t + (2p_i-1) \varepsilon \bv, \varepsilon \geq 0.
\]
Hence, with
\[
\bbeta\in\R^n,\ \beta_i := (2p_i-1) \varepsilon \sqrt{a_i},
\]
we obtain the rank-$\leq 2$ decomposition
\begin{equation}\label{eq:B-decomp}
\cA(\bP) = \bu \bgamma^\top + \bv \bbeta^\top.
\end{equation}
We claim that, for any $\bbeta\neq 0$,
\begin{equation}\label{eq:nuclear-strict}
\|\bu \bgamma^\top + \bv \bbeta^\top\|_{S_1} > \|\bu \bgamma^\top\|_{S_1} = \|\bgamma\|.
\end{equation}
Indeed, let $\bQ\in\R^{d\times d}$ be an orthogonal matrix whose first two columns are $\bu$ and $\bv$. Orthogonal invariance of singular values implies
\[
\|\bu \bgamma^\top + \bv \bbeta^\top\|_{S_1}
=\|\begin{bmatrix}\bgamma & \bbeta & 0 & \cdots & 0\end{bmatrix}\|_{S_1}
=\sigma_1+\sigma_2,
\]
where $\sigma_1\ge\sigma_2\ge 0$.
If $\bbeta$ is not colinear with $\bgamma$, the matrix has rank $2$, so $\sigma_2>0$, and $\sigma_1\geq \|\bgamma\|_2$ (since $\|\bM\|_{2}\ge$ the Euclidean norm of any row). Hence $\sigma_1+\sigma_2>\|\bgamma\|_2$.
If instead $\bbeta=c \bgamma$ for some $c\neq 0$, then the matrix has rank $1$ with singular value $\sqrt{\|\bgamma\|_2^2 + \|\bbeta\|_2^2}=\sqrt{1+c^2} \|\bgamma\|_2>\|\bgamma\|_2$.
Thus~\eqref{eq:nuclear-strict} holds in all cases $\beta\neq 0$.

Combining~\eqref{eq:fund-lb} and~\eqref{eq:within-cluster-Szero}, for any within-cluster feasible $P$,
\begin{equation}\label{eq:obj-diff-within}
F_\lambda(\bP)-F_\lambda \left(\bP^{\star}\right)
 \geq \lambda( \|\cA(\bP)\|_{S_1}-\|\bB^\star\|_{S_1} )
 = \lambda ( \|\bu \bgamma^\top + \bv \bbeta^\top\|_{S_1}-\|\bgamma\|_2 ),
\end{equation}
which is strictly positive by~\eqref{eq:nuclear-strict} whenever $\beta\neq 0$, i.e., whenever some $p_i\neq \tfrac12$.

\medskip
\noindent\textbf{Step 4: Optimality and uniqueness for $\lambda\in[0,\Lambda-2\rho)$.}
Let $\lambda\in[0,\Lambda-2\rho)$. For any feasible $\bP$, decompose $\bP-\bP^{\star}$ into an across-cluster part and a within-cluster part. By~\eqref{eq:reduced-cost-gap},
\[
\langle S(\lambda, \bG^\star), \bP-\bP^{\star}\rangle > 0
\]
if $\bP$ sends any mass across clusters, and~\eqref{eq:fund-lb} implies $F_\lambda(\bP)>F_\lambda (\bP^{\star})$ in that case.
Therefore, any minimizer of $F_\lambda$ must be supported within clusters. For within-cluster couplings, \eqref{eq:obj-diff-within} implies $F_\lambda(\bP)>F_\lambda (\bP^{\star})$ unless $p_i=\tfrac12$ for all $i$, i.e., if $\bP = \bP^\star$.
Consequently, $\bP^{\star}$ is the \emph{unique} minimizer of the Schatten OT problem for $\lambda\in[0,\Lambda-2\rho)$.
\end{proof}

\section{The Gaussian Case}
\label{sec:gaussian}

The previous section treated recovery of low-rank structures in discrete OT. We now discuss an application of the continuous Schatten regularization \eqref{eq:schatten_ot_cts} for Gaussians.

We treat two Gaussian specializations of the Schatten-$p$ programs discussed earlier:
(i) a nuclear-norm penalty that promotes low-rank cross-covariance, and
(ii) a nuclear-norm penalty that promotes low-rank transport.
As emphasized in our general framework, the barycentric projection $x\mapsto \mathbb{E}_\pi[Y\mid X=x]$
is an \emph{affine} map of the coupling $\pi$, so the induced Schatten-$p$ penalty is convex in~$\pi$ for $p \geq 1$. The same
holds for Schatten penalties applied to any affine image $A(\pi)$.

For simplicity, we consider the mean zero case.
Let $\mu=\mathcal{N}(\bzero,\bSigma_0)$ and $\nu=\mathcal{N}(\bzero,\bSigma_1)$ on $\mathbb{R}^d$ with
$\bSigma_0,\bSigma_1\succ0$. A \emph{Gaussian coupling} is a joint Gaussian
$\pi=\mathcal{N} \big(\begin{bmatrix}\bzero\\ \bzero\end{bmatrix}, \begin{bmatrix}\bSigma_0 & \bK\\ \bK^\top & \bSigma_1\end{bmatrix}\big)$,
parameterized by a cross-covariance $\bK\in\mathbb{R}^{d\times d}$ satisfying the feasibility constraint
\begin{equation}
\label{eq:block-PSD}
\begin{bmatrix}\bSigma_0 & \bK\\ \bK^\top & \bSigma_1\end{bmatrix}\succeq 0
\iff
\|\bSigma_0^{-1/2} \bK \bSigma_1^{-1/2}\|_2\leq 1.
\end{equation}
We denote the set of all such $\bK$ as $\cK(\bSigma_0, \bSigma_1)$. Equivalently, we denote the set of all Gaussian couplings between $\mu$ and $\nu$ as $\Pi_g(\mu, \nu)$.

For the quadratic cost $c(\bx,\by)=\|\bx-\by\|^2$, the transport cost under $\pi$ is
$\mathbb{E}_\pi\|X-Y\|^2=\mathrm{tr}(\bSigma_0)+\mathrm{tr}(\bSigma_1)-2 \mathrm{tr}(\bK)$.
Moreover the barycentric map induced by $\pi$ is
\begin{equation}
\label{eq:bary-map}
T_\pi(\bx)= \bA_\pi\bx,\ \bA_\pi:=\bK^\top\bSigma_0^{-1}.
\end{equation}

\subsection{Gaussian Low-rank Cross-Covariance}

Consider the Gaussian Schatten OT problem
\begin{equation}
\label{eq:K-problem}
\min_{\bK \in \cK(\bSigma_0, \bSigma_1)}
  \mathrm{tr}(\bSigma_0)+\mathrm{tr}(\bSigma_1)-2 \mathrm{tr}(\bK)
 + \lambda\|\bK\|_{S_1}.
\end{equation}
This is a semidefinite program.

We can solve this problem in closed form. Let $\bS:=\bSigma_1^{1/2}\bSigma_0^{1/2}$ and write its SVD $\bS=\bU\diag(\sigma_1,\dots,\sigma_d)\bV^\top$ with
$\sigma_1\geq\cdots\ge\sigma_d>0$. Feasible $\bK$ can be written as $\bK=\bSigma_0^{1/2}\bM\bSigma_1^{1/2}$ with
$\|\bM\|_2\leq 1$. By von Neumann’s trace inequality, $\mathrm{tr}(\bK)=\mathrm{tr}(\bS\bM)\leq \sum_i \sigma_i s_i$
where $s_i$ are the singular values of $\bM$, with equality when $\bM$ shares singular vectors with $\bS$.
At optimum we may thus take $\bM=\bU\diag(s_1,\dots,s_d)\bV^\top$ with $0\le s_i\le 1$ and
$\|\bK\|_{S_1}=\|\bM\|_{S_1}=\sum_i s_i$. The objective in \eqref{eq:K-problem} reduces (up to constants) to
\begin{equation*}
\min_{0\le s_i\le 1} \sum_{i=1}^d \big(\lambda-2\sigma_i\big)s_i
  = \sum_{i=1}^d \min_{0\le s\le 1} (\lambda-2\sigma_i)s
  ,
\end{equation*}
which is separable and linear in each $s_i$. Hence, we arrive at the following proposition.

\begin{proposition}[Hard spectral selection]
\label{prop:K-hard-thresh}
The unique minimizer of \eqref{eq:K-problem} is obtained by \emph{hard thresholding} the singular spectrum of $\bS$:
\begin{equation}
\label{eq:K-solution}
s_i^\star=\mathbf{1}\{\sigma_i>\lambda/2\},\qquad
\bK_ \lambda=\bSigma_0^{1/2} \bU\diag(s_i^\star) \bV^\top \bSigma_1^{1/2}.
\end{equation}
In particular, $\operatorname{rank}(\bK_ \lambda)=\#\{i:\sigma_i>\lambda/2\}$.
\end{proposition}

We include examples to illustrate this hard-thresholding rule.
\begin{itemize}
    \item[(i)] \emph{Isotropic case:} $\bSigma_0=a^2 \bI, \bSigma_1=b^2 \bI$ gives $\bS=ab \bI$ and hence either $\bK_ \lambda=ab \bI$ if $\lambda<2ab$, or $\bK_ \lambda=0$ if $\lambda\geq 2ab$.
    \item[(ii)] \emph{Commuting covariances:} If $\bSigma_0=\bU\diag(\ba)\bU^\top$ and $\bSigma_1= \bU\diag(\bb)\bU^\top$, then $\sigma_i=\sqrt{a_i b_i}$ and $\bK_ \lambda= \bU\diag\big(\sqrt{\ba \odot \bb} \mathbf{1}\{\sqrt{\ba \odot \bb}>\lambda/2\}\big)\bU^\top$. Here, $\odot$ is the Hadamard (elementwise) product.
\end{itemize}
We note that the inclusion of a Schatten-2 penalty in this program results in soft thresholding.

\subsection{Gaussian Barycentric Displacements}

We now penalize the (weighted) barycentric displacement. In this case, the resulting convex program is
\begin{equation}
\label{eq:disp-problem}
\min_{\pi \in \Pi_g(\mu, \nu)}
  \mathrm{tr}(\bSigma_0)+\mathrm{tr}(\bSigma_1)-2 \mathrm{tr}(\bK) + \lambda \big\| (\bA_\pi-\bI)\bSigma_0^{1/2}\big\|_{S_1},
\end{equation}
with $\bA_\pi$ as in \eqref{eq:bary-map}.

We can give a closed form when $\bSigma_0$ and $\bSigma_1$ commute, which already reveals a clear structure. Suppose there exists an orthogonal $\bU$ with
$\bSigma_0=\bU\diag(\ba)\bU^\top$ and
$\bSigma_1=\bU\diag(\bb)\bU^\top$. Any feasible $\bK$ aligned with $\bU$ takes the form
$\bK=\bU\diag(\sqrt{\ba \odot \bb} \odot \bm )U^\top$ with $0\leq m_i\leq 1$. Then $\bA_\pi=\bK^\top\bSigma_0^{-1}$ is diagonal in the same basis with entries
\[
\alpha_i:=\frac{\sqrt{a_i} m_i \sqrt{b_i}}{a_i}=m_i \sqrt{\frac{b_i}{a_i}},
\]
which implies that the diagonal values of $(\bA_\pi-I)\bSigma_0^{1/2}$ are $ m_i\sqrt{b_i}-\sqrt{a_i}$ in this basis as well.
Hence $\|(\bA_\pi-\bI)\bSigma_0^{1/2}\|_{S_1}=\sum_i |m_i\sqrt{b_i}-\sqrt{a_i}|$, and
\[
\mathrm{tr}(\bK)=\sum_i \sqrt{a_i} m_i \sqrt{b_i}.
\]
Consequently, \eqref{eq:disp-problem} decouples into $d$ scalar problems over $m_i\in[0,1]$:
\begin{equation}
\label{eq:scalar-disp}
\min_{0\le m\le1} \phi_{\ba, \bb,\lambda}(m):=
-2\sqrt{ab} m + \lambda |m\sqrt{b}-\sqrt{a}|.
\end{equation}

\begin{theorem}
\label{thm:disp-solution}
Fix $(\ba, \bb,\lambda)$ with $\ba, \bb>0$. The unique minimizer of \eqref{eq:scalar-disp} is
\[
m^\star=
\begin{cases}
1, & \text{if } b\le a, \\[4pt]
1, & \text{if } b>a \text{ and } \lambda<2\sqrt{a},\\[4pt]
\sqrt{a/b}, & \text{if } b>a \text{ and } \lambda\ge 2\sqrt{a}.
\end{cases}
\]
Equivalently, the coupling that solves \eqref{eq:disp-problem} has barycentric projection map $T_{\pi}(\cdot) = \bA_{\lambda} \cdot$, where
\begin{equation}
\label{eq:A-solution}
\bA_\lambda = \bU \diag (\balpha^\star) \bU^\top ,
\qquad
\alpha_i^\star=
\begin{cases}
\sqrt{b_i/a_i}, & \text{if } b_i\le a_i,\\[2pt]
\sqrt{b_i/a_i}, & \text{if } b_i>a_i \text{ and } \lambda<2\sqrt{a_i},\\[2pt]
1, & \text{if } b_i>a_i \text{ and } \lambda\ge 2\sqrt{a_i}.
\end{cases}
\end{equation}
\end{theorem}

The proof follows from analyzing the 1-dimensional optimization problem \eqref{eq:scalar-disp}.

In words, the regularizer does not suppress contracting directions ($b_i\le a_i$), and it
prunes expanding directions $b_i>a_i$ back to the identity once $\lambda$ crosses the sharp threshold $2\sqrt{a_i}$. Thus, $\operatorname{rank}(\bA_\lambda-\bI)$ equals the number of contracting eigendirections plus the number of expanding eigendirections with $\lambda<2\sqrt{a_i}$.

As an example, consider the isotropic case. If $\bSigma_0=\sigma_0^2 \bI, \bSigma_1=\sigma_1^2 \bI$.
If $\sigma_1\le\sigma_0$, then $\bA_\lambda=(\sigma_1/\sigma_0)\bI$ for all $\lambda$. On the other hand, if $\sigma_1>\sigma_0$, then $\bA_\lambda=(\sigma_1/\sigma_0)\bI$ for $\lambda<2\sigma_0$,
and $\bA_\lambda=I$ for $\lambda\geq 2\sigma_0$. 

%




\subsection{Discussion}

In the Gaussian case, for $p>1$, the programs remain convex. In the commuting case, we can find separable one-dimensional convex problems similar to those we found in the previous sections. For larger $p$, we would observe smooth shrinkage of the spectrum $m_i$ rather than hard thresholds at $p=1$. We leave a detailed analysis of the noncommuting case to future work.

\section{Details on 4i Experiment Setup}
\label{app:exp_details}

For reproducibility, we give more details on the 4i experiment here. We subsample source and target points to form measures with 1000 points each. We then run mirror descent with the desired regularization, where the initial step size is $\eta_0 = 0.1$ for low-rank coupling recovery and $\eta_0 = 10^{-4}$ for low-rank barycentric map recovery. A diminishing step size of $\eta_k = \eta_0 / \sqrt{k}$ is used. We run mirror descent for a maximum of 50 iterations, with the Sinkhorn projection run for 500 iterations at each iteration, unless the marginal error reaches $10^{-12}$. This is followed by the rounding procedure of \cite{altschuler2017near} to ensure iterates remain in the transport polytope. We return the averaged iterate as our proposed solution to the Schatten OT problem.

\end{document}